\documentclass{article}

\usepackage[final]{neurips_2025}

\usepackage[utf8]{inputenc}
\usepackage[T1]{fontenc}
\usepackage{amsmath,amssymb,amsfonts,amsthm}
\usepackage{mathtools}
\usepackage{bm}
\usepackage{booktabs}
\usepackage{graphicx}
\graphicspath{{figures/}}
\usepackage[table]{xcolor}
\usepackage{hyperref}
\usepackage{tikz}
\usepackage{subcaption}
\usepackage{wrapfig}
\usepackage{float}
\usepackage[above,below]{placeins}
\usetikzlibrary{arrows.meta, positioning, calc, shapes.geometric, decorations.pathreplacing, patterns, fit, matrix}

\newtheorem{theorem}{Theorem}
\newtheorem{proposition}[theorem]{Proposition}

\DeclareMathOperator{\diag}{diag}
\DeclareMathOperator{\softplus}{softplus}
\DeclareMathOperator{\softmax}{softmax}
\DeclareMathOperator{\ReLU}{ReLU}

\newcommand{\R}{\mathbb{R}}
\newcommand{\Lcal}{\mathcal{L}}
\newcommand{\Ecal}{\mathcal{E}}

\newcommand{\Ncal}{\mathcal{N}}

\newcommand{\psifield}{\bm{\psi}}

\title{Metriplector: From Field Theory to Neural Architecture}

\author{
  Dan Oprisa$^{*}$ \quad
  Peter Toth$^{*}$ \\[4pt]
  Spheroid Labs\thanks{\href{https://spheroid.ai}{spheroid.ai}. $^{*}$Equal contribution. Correspondence: \{dan, peter\}@spheroid.ai}
}

\begin{document}

\maketitle

\begin{abstract}
We present \textbf{Metriplector}, a neural architecture primitive in which the input configures an abstract physical system---fields, sources, and operators---and the dynamics of that system \emph{is} the computation.
Multiple fields evolve via coupled metriplectic dynamics, and the stress-energy tensor $T^{\mu\nu}$, derived from Noether's theorem, provides the readout.
The metriplectic formulation admits a natural spectrum of instantiations: the dissipative branch alone yields a screened Poisson equation solved exactly via conjugate gradient; activating the full structure---including the antisymmetric
Poisson bracket---gives field dynamics for image recognition, language modeling, and robotic control.
We evaluate Metriplector across five domains, each using a task-specific architecture built from this shared primitive with progressively richer physics: \textbf{81.03\%} on CIFAR-100 with 2.26M parameters; \textbf{88\% CEM success} on Reacher robotic control with under 1M parameters; \textbf{97.2\%} exact Sudoku solve rate with zero structural injection; \textbf{1.182 bits/byte} on language modeling with 3.6$\times$ fewer training tokens than a GPT baseline; and \textbf{F1\,=\,1.0} on maze pathfinding, generalizing from 15$\times$15 training grids to unseen 39$\times$39 grids.
\end{abstract}

\section{Introduction}
\label{sec:intro}

Physics offers a rich source of principled structures for computation:
conservation laws that guarantee what information is preserved, variational
principles that determine how systems evolve, and symmetry theorems that
connect structure to observables.
These are not heuristics---they are exact, composable, and universal.
If they can be instantiated as trainable architecture primitives, the
physical structure itself carries part of the computational load,
potentially reducing the capacity needed to solve a problem.
The long-term vision is a \emph{standard model} of neural
computation---a unified framework of physics-grounded primitives, each
derived from a different physical principle, that can be composed the way
physical theories are composed to describe nature.

This paper takes a step in that direction: \emph{the energy landscape is the program}.
The input to a computation configures an abstract physical system---its boundary conditions, source terms, conductances, and advection operators.
The dynamics of that system \emph{is} the computation: fields $\psifield$ evolve under the metriplectic (GENERIC) equation, and the stress-energy tensor $T^{\mu\nu}$---derived from Noether's theorem---provides the readout.
A central finding is that what appear to be domain-specific architectures for recognition (CIFAR-100), reasoning (Sudoku), and robotic control (Reacher) are in fact \emph{the same physics}: both read off $T^{\mu\nu}$ from evolved fields, and the multigrid object layer that discovers Sudoku boxes performs learned coarse-graining analogous to block-spin methods~\citep{kadanoff1966scaling}.

\begin{figure}[!htb]
\centering
\includegraphics[width=\textwidth]{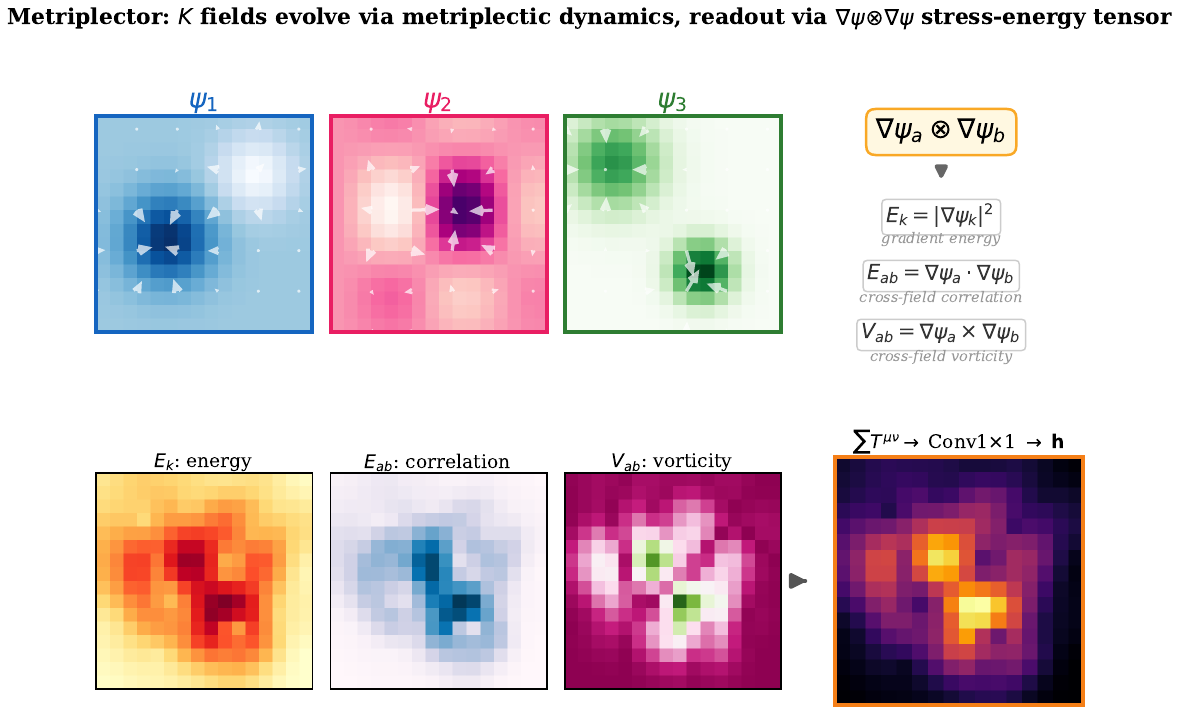}
\caption{\textbf{Metriplector field interaction.} \emph{Top:} $K$ fields $\psifield_k$ evolve via metriplectic dynamics over the spatial grid (gradient arrows show $\nabla\psifield_k$); the outer product $\nabla\psifield_a \otimes \nabla\psifield_b$ yields three stress-energy components. \emph{Bottom:} per-field gradient energy $E_k = |\nabla\psifield_k|^2$, cross-field correlation $E_{ab} = \nabla\psifield_a \cdot \nabla\psifield_b$, and vorticity $V_{ab} = \nabla\psifield_a \times \nabla\psifield_b$, summed and projected via Conv$1{\times}1$ into $\mathbf{h}$. Shown for $K{=}3$; the full model uses $K{=}32$.}
\label{fig:field_schematic}
\end{figure}

We develop this idea from first principles.
We begin with the physics: Lagrangian and Hamiltonian mechanics, the GENERIC framework~\citep{grmela1997dynamics,ottinger2005beyond} that unifies reversible and irreversible dynamics in a single equation, and Noether's theorem~\citep{noether1918invariante} connecting symmetries to conservation laws (Section~\ref{sec:physics}).
We then show how residual neural network layers can be understood as dynamics on energy landscapes, and how metriplectic structure provides conservation guarantees that general-purpose architectures lack.

From this foundation, we distill \textbf{Metriplector}: a physics-native architecture primitive that instantiates the metriplectic equation differently for each domain (Section~\ref{sec:method}).
In its simplest form, the metriplectic equation reduces to a screened Poisson equation on a graph:
\begin{equation}
    (L_W + \Lambda)\psifield = \mathbf{b},
    \label{eq:core}
\end{equation}
where $L_W$ is a learned weighted graph Laplacian, $\Lambda$ is per-node damping, and $\mathbf{b}$ is a learned source vector.
For spatial reasoning, this is solved exactly via conjugate gradient; for language modeling, the causal structure allows $O(N \log N)$ parallel scan solutions; for recognition, the full metriplectic dynamics---including the antisymmetric Poisson bracket---is evolved via Euler integration.

\paragraph{Contributions.}
\begin{enumerate}
    \item A \textbf{physics-native architecture primitive} grounded in the
    metriplectic framework, in which the input configures an abstract
    physical system and the dynamics is the computation
    (Section~\ref{sec:physics}).

    \item The \textbf{stress-energy tensor $T^{\mu\nu}$ as a principled
    readout}: we show that the conserved quantity from Noether's theorem
    provides the most effective feature extraction from evolved fields,
    outperforming heuristic alternatives by 0.6--2.6 accuracy points on
    CIFAR-100 (Section~\ref{sec:cifar100_method}). The same readout
    (diagonal $T^{\mu\nu}$) underlies the Sudoku instantiation, unifying
    recognition and constraint satisfaction under a single physics
    (Section~\ref{sec:discussion}).

    \item A \textbf{metriplectic spectrum} of instantiations with
    progressively richer dynamics: the dissipative branch alone (screened
    Poisson, solved exactly via CG) suffices for spatial reasoning; the
    full metriplectic structure (Poisson bracket + metric tensor) is
    needed for recognition and world modeling; and a causal variant enables language modeling
    (Section~\ref{sec:method}).

    \item Two \textbf{transferable design principles} validated by
    ablation: the \emph{operator-from-input} separation, where the
    representation $\mathbf{h}$ defines the physics and the fields
    $\psifield$ are the solution (removing it costs 14.3 accuracy points);
    and \emph{zero structural injection}, where task structure is
    discovered from data rather than encoded in the graph topology
    (Section~\ref{sec:discussion}).

    \item \textbf{Cross-domain experimental validation}: 81.03\% on
    CIFAR-100 (2.26M params); 97.2\% exact Sudoku solve (120K params,
    zero injection); 1.182 BPB on language modeling (3.6$\times$ fewer
    tokens than GPT); F1\,=\,1.0 on maze pathfinding (43.8K params) with
    generalization to unseen 39$\times$39 grids; and 88\% CEM success
    on Reacher robotic control (913K params), competitive with the
    strongest baseline at 16$\times$ fewer parameters
    (Section~\ref{sec:experiments}).

    \item A \textbf{physics-native world model} for robotic control
    that instantiates metriplectic dynamics as a latent-space predictor
    with symplectic Poisson bracket, port-Hamiltonian action conditioning,
    and multi-step autoregressive training, achieving 88\% CEM
    success with 16$\times$ fewer parameters than LeWM (86\%, $\sim$15M)
    (Section~\ref{sec:reacher_method}).
\end{enumerate}

\section{From Physics to Neural Computation}
\label{sec:physics}

\subsection{From Lagrangians to Hamiltonians}
\label{sec:lagrangian_to_hamiltonian}

The foundation of all physics is the \emph{principle of stationary action}.
A physical system with generalised coordinates $q$ and velocities $\dot{q}$ is described by a Lagrangian $\mathcal{L}(q, \dot{q}, t) = T - V$, where $T$ is kinetic and $V$ potential energy.
The true trajectory extremises the action functional:
\begin{equation}
    \mathcal{A}[q] = \int_{t_0}^{t_1} \mathcal{L}(q, \dot{q}, t)\, dt, \qquad
    \delta \mathcal{A} = 0 \;\implies\; \frac{d}{dt}\frac{\partial \mathcal{L}}{\partial \dot{q}_i} - \frac{\partial \mathcal{L}}{\partial q_i} = 0.
    \label{eq:euler_lagrange}
\end{equation}
These are the Euler--Lagrange equations---the most general statement of classical dynamics.

The \emph{Hamiltonian formulation} reformulates this in terms of positions $q$ and conjugate momenta $p_i = \partial \mathcal{L}/\partial \dot{q}_i$:
\begin{equation}
    H(q, p) = \sum_i p_i \dot{q}_i - \mathcal{L}, \qquad
    \dot{q}_i = \frac{\partial H}{\partial p_i}, \quad
    \dot{p}_i = -\frac{\partial H}{\partial q_i}.
    \label{eq:hamilton_eqs}
\end{equation}
Hamilton's equations are equivalent to the Euler--Lagrange equations but reveal a deeper structure: the \emph{symplectic structure} of phase space.
The state $z = (q, p)$ evolves as $\dot{z} = J_0 \nabla H$, where $J_0 = \bigl(\begin{smallmatrix} 0 & I \\ -I & 0 \end{smallmatrix}\bigr)$ is the canonical symplectic matrix---skew-symmetric by construction, guaranteeing exact energy conservation:
\begin{equation}
    \frac{dH}{dt} = \nabla H^\top J_0 \nabla H = 0.
    \label{eq:energy_conservation}
\end{equation}
This is the key insight for neural architecture design: \emph{the dynamics is determined entirely by the energy landscape $H$ and the structure matrix $J_0$}.
The energy landscape encodes \emph{what} the system knows; the structure matrix encodes \emph{how} information flows.
Neural networks that respect this separation inherit the stability, conservation, and interpretability of Hamiltonian mechanics.

\paragraph{From particles to fields.}
When the degrees of freedom are continuous---a vibrating membrane, an electromagnetic field, a neural feature map---the Lagrangian becomes a functional over fields $\psifield(x, t)$:
\begin{equation}
    \mathcal{A}[\psifield] = \int \mathcal{L}\!\left(\psifield, \nabla\psifield, \partial_t\psifield\right) d^n x\, dt.
    \label{eq:field_action}
\end{equation}
The Euler--Lagrange equations become partial differential equations.
A 2D convolutional feature map at layer $\ell$ is precisely such a field: $\psifield^{(\ell)}(x, y)$ is a function on a spatial grid, and the convolution kernel defines the local interaction Lagrangian.
Metriplector makes this analogy literal: it evolves $K$ physics fields $\psifield_1, \ldots, \psifield_K$ on a spatial grid under dynamics derived from a Hamiltonian energy functional.

\subsection{Every Physical System in One Equation}
\label{sec:generic_framework}

The GENERIC (General Equation for Non-Equilibrium Reversible-Irreversible Coupling) framework~\citep{grmela1997dynamics,ottinger2005beyond} unifies all of classical physics:
\begin{equation}
    \dot{z} = \underbrace{L(z) \cdot \nabla E(z)}_{\text{reversible (Hamiltonian)}} + \underbrace{M(z) \cdot \nabla S(z)}_{\text{irreversible (dissipative)}},
    \label{eq:generic}
\end{equation}
with degeneracy conditions:
\begin{align}
    M \cdot \nabla E &= 0 \quad \text{(dissipation cannot change energy)}, \nonumber \\
    L \cdot \nabla S &= 0 \quad \text{(Hamiltonian dynamics cannot produce entropy)}.
    \label{eq:degeneracy}
\end{align}

The Hamiltonian channel ($L \cdot \nabla E$) describes reversible dynamics: $L$ is skew-symmetric, so $dE/dt = \nabla E^\top L \nabla E = 0$---energy is exactly conserved.
The dissipative channel ($M \cdot \nabla S$) describes irreversible dynamics: $M$ is symmetric positive semi-definite, so $dS/dt = \nabla S^\top M \nabla S \geq 0$---entropy never decreases.
Together, these structural properties are \emph{built into the architecture}, not learned from data.

\subsection{Metriplector as Dissipative GENERIC}

Each Poisson solve in Metriplector finds the steady state of the dissipative branch ($L = 0$):
\begin{equation}
    0 = M \cdot \nabla S(\psifield) \implies (L_W + \Lambda)\psifield = \mathbf{b},
\end{equation}
where the Onsager matrix $M = L_W + \Lambda$ is SPD (satisfying the Onsager reciprocal relations~\citep{onsager1931reciprocal}) and the entropy functional is $S(\psifield) = -\frac{1}{2}\psifield^\top(L_W + \Lambda)\psifield + \mathbf{b}^\top\psifield$.

The Poisson solution minimizes the screened Dirichlet energy:
\begin{equation}
    \Ecal_\text{Dir}(\psifield) = \frac{1}{2}\sum_{(i,j) \in E} w_{ij}(\psifield_i - \psifield_j)^2 + \frac{1}{2}\sum_i \Lambda_i \psifield_i^2 - \sum_i b_i \psifield_i.
    \label{eq:dirichlet}
\end{equation}
Since the Hessian $L_W + \Lambda$ is SPD (Proposition~\ref{prop:wellposed}), the critical point is the unique global minimum---the physical equilibrium.

\subsection{Full Metriplectic Dynamics for Recognition}
\label{sec:full_metriplectic}

For reasoning tasks (maze, Sudoku), the dissipative branch alone suffices: the equilibrium of the screened Poisson equation encodes the answer, and the Hamiltonian channel plays no role ($L = 0$).
For recognition (CIFAR-100), both branches of the GENERIC equation~\eqref{eq:generic} are active:
\begin{equation}
    \psifield \leftarrow \psifield + \Delta t \left[\underbrace{-\sigma \cdot \nabla^2 \psifield}_{\text{diffusion } (M)} + \underbrace{\alpha \cdot J_{\text{anti}}\,\psifield}_{\text{advection } (L)} - \underbrace{\gamma \psifield}_{\text{damping}} + \underbrace{\mathbf{s}}_{\text{source}}\right],
    \label{eq:euler_metriplectic}
\end{equation}
where $J$ is a raw learned $K{\times}K$ matrix and $J_{\text{anti}} = J - J^\top$ is its antisymmetrization, yielding a skew-symmetric Poisson tensor that mediates cross-field coupling.
Throughout the paper, references to ``$J$'' in the context of the Poisson bracket refer to this antisymmetrized form $J_{\text{anti}}$ unless otherwise noted.
$\nabla^2$ is a depthwise convolution approximating the Laplacian on the image grid, and the coefficients $\sigma, \alpha, \gamma, \mathbf{s}$ are produced from the representation~$\mathbf{h}$ via Conv$1{\times}1$ projections.

The Poisson bracket $J_{\text{anti}}$ enables \emph{advection}---it rotates information between the $K$ fields without dissipating it.
The metric tensor $M$ (diffusion + damping) enables \emph{spatial smoothing}---spreading information across the image grid and damping irrelevant variation.
Recognition requires both: spatial smoothing builds invariance to local perturbations, while cross-field advection constructs the feature interactions needed for fine-grained class discrimination.

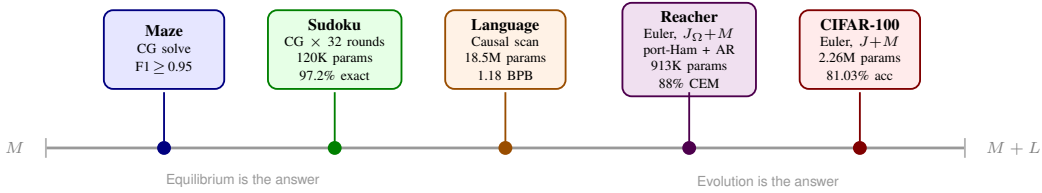
\begin{figure}[ht]
\centering
\resizebox{\textwidth}{!}{%
\begin{tikzpicture}[
    dombox/.style={draw, thick, rounded corners=4pt, minimum height=1.2cm, minimum width=1.8cm, align=center, font=\scriptsize},
    arr/.style={-{Stealth[length=2mm]}, very thick, black!30},
]
\draw[very thick, black!40] (0, 0) -- (14, 0);
\draw[thick, black!30] (0, -0.15) -- (0, 0.15);
\draw[thick, black!30] (14, -0.15) -- (14, 0.15);
\node[font=\scriptsize\sffamily\bfseries, black!50, anchor=east] at (-0.2, 0) {$M$};
\node[font=\scriptsize\sffamily\bfseries, black!50, anchor=west] at (14.2, 0) {$M + L$};

\node[dombox, fill=blue!10, draw=blue!50!black, text width=1.6cm] (maze) at (1.8, 1.5) {\centering\textbf{Maze}\\\tiny CG solve\\F1\,$\geq$\,0.95};
\draw[thick, blue!50!black] (1.8, 0.1) -- (1.8, 0.85);
\fill[blue!50!black] (1.8, 0) circle (3pt);

\node[dombox, fill=green!10, draw=green!50!black, text width=1.8cm] (sudoku) at (4.4, 1.5) {\centering\textbf{Sudoku}\\\tiny CG $\times$ 32 rounds\\120K params\\97.2\% exact};
\draw[thick, green!50!black] (4.4, 0.1) -- (4.4, 0.85);
\fill[green!50!black] (4.4, 0) circle (3pt);

\node[dombox, fill=orange!10, draw=orange!60!black, text width=1.6cm] (lang) at (7.0, 1.5) {\centering\textbf{Language}\\\tiny Causal scan\\18.5M params\\1.18 BPB};
\draw[thick, orange!60!black] (7.0, 0.1) -- (7.0, 0.85);
\fill[orange!60!black] (7.0, 0) circle (3pt);

\node[dombox, fill=violet!12, draw=violet!60!black, text width=1.8cm] (reacher) at (9.8, 1.5) {\centering\textbf{Reacher}\\\tiny Euler, $J_\Omega{+}M$\\port-Ham + AR\\913K params\\88\% CEM};
\draw[thick, violet!60!black] (9.8, 0.1) -- (9.8, 0.85);
\fill[violet!60!black] (9.8, 0) circle (3pt);

\node[dombox, fill=red!8, draw=red!50!black, text width=1.6cm] (cifar) at (12.4, 1.5) {\centering\textbf{CIFAR-100}\\\tiny Euler, $J{+}M$\\2.26M params\\81.03\% acc};
\draw[thick, red!50!black] (12.4, 0.1) -- (12.4, 0.85);
\fill[red!50!black] (12.4, 0) circle (3pt);

\node[font=\tiny\sffamily, black!40] at (3.0, -0.5) {Equilibrium is the answer};
\node[font=\tiny\sffamily, black!40] at (11.0, -0.5) {Evolution is the answer};

\end{tikzpicture}%
}
\caption{\textbf{The metriplectic spectrum.} All five domains instantiate the same GENERIC equation. Maze and Sudoku use only the dissipative branch ($M$), solved at equilibrium via CG. Language uses causal dissipation via scan. CIFAR-100 and Reacher activate full metriplectic structure via Euler integration; Reacher additionally incorporates canonical symplectic $J_\Omega$, port-Hamiltonian action conditioning, and multi-step AR training.}
\label{fig:spectrum}
\end{figure}

This creates a spectrum of metriplectic instantiations (Figure~\ref{fig:spectrum}):
\begin{center}
\small
\begin{tabular}{lccc}
\toprule
\textbf{Domain} & \textbf{Active channels} & \textbf{Solver} & \textbf{Dynamics} \\
\midrule
Maze & $M$ only & Exact CG & Equilibrium \\
Sudoku & $M$ only & Exact CG & Iterated equilibrium \\
CIFAR-100 & $M + L$ & Euler steps & Progressive evolution \\
Language & $M$ only (causal) & Parallel scan & Causal propagation \\
Reacher & $M + L$ & Euler steps & Action-conditioned evolution \\
\bottomrule
\end{tabular}
\end{center}

\subsection{Symmetries and Conservation Laws}
\label{sec:noether}

Noether's theorem~\citep{noether1918invariante} connects symmetries to conserved currents: every continuous symmetry of the action $\mathcal{A}$ yields a conserved current.

\begin{theorem}[Noether's theorem, 1918]
\label{thm:noether}
If the action $\mathcal{A}[\psifield] = \int \mathcal{L}(\psifield, \partial_\mu\psifield)\, d^n x$ is invariant under a continuous one-parameter family of transformations $\psifield \to \psifield + \epsilon\, \delta\psifield$, then the current
\begin{equation}
    j^\mu = \frac{\partial \mathcal{L}}{\partial (\partial_\mu \psifield_a)} \delta\psifield_a - \Bigl(\frac{\partial \mathcal{L}}{\partial (\partial_\mu \psifield_a)} \partial_\nu\psifield_a - \delta^\mu_\nu \mathcal{L}\Bigr)\delta x^\nu
    \label{eq:noether_current}
\end{equation}
is conserved: $\partial_\mu j^\mu = 0$.
\end{theorem}

For Metriplector, Noether's theorem provides both the theoretical justification for the architecture and the optimal readout features:

\begin{center}
\small
\begin{tabular}{lll}
\toprule
\textbf{Symmetry} & \textbf{Conserved quantity} & \textbf{Role in Metriplector} \\
\midrule
Spatial translation & Stress-energy tensor $T_{\mu\nu}$ & Readout features ($\nabla\psifield \times \nabla\psifield$) \\
Spatial rotation & Angular momentum $L_z$ & Noether readout ($x\, p_y - y\, p_x$) \\
Scale invariance & Dilation current $D$ & Noether readout ($x\, p_x + y\, p_y$) \\
Time translation & Energy $E$ & Conserved by Poisson bracket $J$ \\
Phase rotation & Charge / particle number & Conserved charges of $J$ \\
\bottomrule
\end{tabular}
\end{center}

The stress-energy tensor deserves special attention.
For a system of $K$ fields $\psifield_1, \ldots, \psifield_K$ with Lagrangian density $\mathcal{L} = \frac{1}{2}\sum_a |\nabla\psifield_a|^2 - V(\psifield)$, spatial translation invariance yields:
\begin{equation}
    T_{ab}^{ij} = \partial_i \psifield_a \cdot \partial_j \psifield_b, \qquad i, j \in \{x, y\}, \quad a, b \in \{1, \ldots, K\}.
    \label{eq:stress_energy_physics}
\end{equation}
The diagonal terms $T_{aa}^{ii}$ are the energy density of field $a$ in direction $i$; the off-diagonal terms encode cross-field correlations.
This is precisely the readout that Metriplector uses for CIFAR-100 classification (Section~\ref{sec:cifar100_method})---not as a heuristic, but as the \emph{natural observable} dictated by the symmetries of the dynamics.
Section~\ref{sec:cifar100_method} shows that this physics-derived readout yields 0.6--2.6 accuracy points above heuristic alternatives.

\subsection{Algebraic Structure: Lie Groups and the Poisson Bracket}
\label{sec:algebraic_structure}

The Poisson bracket in GENERIC has a deep algebraic origin: it is a Lie bracket on the space of observables~\citep{arnold1989mathematical,marsden1999introduction}.

\paragraph{Lie algebra of observables.}
For any two observables $F, G$ on phase space, the Poisson bracket
\begin{equation}
    \{F, G\} = \sum_i \left(\frac{\partial F}{\partial q_i}\frac{\partial G}{\partial p_i} - \frac{\partial F}{\partial p_i}\frac{\partial G}{\partial q_i}\right) = (\nabla F)^\top J_0\, \nabla G
    \label{eq:poisson_bracket}
\end{equation}
satisfies skew-symmetry ($\{F, G\} = -\{G, F\}$), the Jacobi identity ($\{F, \{G, H\}\} + \text{cyclic} = 0$), and the Leibniz rule.
These are exactly the axioms of a Lie algebra.
By Noether's theorem, every continuous symmetry of the energy functional has a corresponding conserved quantity.
In Metriplector, the learned Poisson tensor $J$ defines a Lie algebra on the $K$-field space.
The skew-symmetry of $J$ guarantees energy conservation ($dH/dt = 0$ along the Hamiltonian flow).
Crucially, the stress-energy tensor $T^{\mu\nu}$ used as the readout mechanism (Section~\ref{sec:method}) is precisely the collection of Noether currents associated with translational symmetry of the learned field dynamics---these conserved quantities \emph{emerge} from the learned energy landscape and change as training progresses.

\paragraph{Symmetric and antisymmetric decomposition.}
Any bilinear interaction between field components $\psifield_a$ and $\psifield_b$ decomposes into a symmetric part and an antisymmetric part.
In the metriplectic framework, these have distinct physical roles:
\begin{itemize}
    \item \textbf{Symmetric} ($M_{ab} = M_{ba}$, positive semi-definite): governs the \emph{metric bracket}---dissipation, diffusion, irreversible relaxation. The Onsager reciprocal relations~\citep{onsager1931reciprocal} require this symmetry.
    \item \textbf{Antisymmetric} ($J_{ab} = -J_{ba}$): governs the \emph{Poisson bracket}---advection, rotation, reversible dynamics. The skew-symmetry guarantees exact energy conservation along the Hamiltonian flow.
\end{itemize}
The metriplectic equation $\dot{z} = (M + L)\nabla E$ decomposes the dynamics into these two complementary channels by construction.
In $K$-dimensional field space, the Poisson bracket $J$ couples all $K$ fields via a learned $K{\times}K$ antisymmetric matrix ($J_{\text{anti}} = J - J^\top$), while the stress-energy readout extracts all $K^2$ pairwise interactions: $K$ diagonal (gradient energy), $K(K{-}1)/2$ symmetric (cross-field correlation), and $K(K{-}1)/2$ antisymmetric (vorticity).
For $K{=}32$, this yields $1{,}024$ physics features---a complete decomposition of the inter-field coupling space.
The $K$-field bottleneck (projecting from $D{=}128$ representation space to $K{=}32$ physics space) makes this completeness tractable.

\subsection{Neural Architectures as Hamiltonian Systems}
\label{sec:hamiltonian_architectures}

Residual neural network layers can be analysed as dynamics on energy landscapes. The differences between architectures reduce to (i) the Hamiltonian $H$, (ii) the dynamics (dissipative, conservative, or both), and (iii) the spatial coupling.

\paragraph{The universal form.}
Any residual layer $\psifield_{\text{new}} = \psifield + \Delta\psifield$ can be analysed by asking: is $\Delta\psifield$ the gradient of some energy function $H(\psifield)$? If $\Delta\psifield = -\nabla H$, the layer performs gradient descent on $H$. If $\Delta\psifield = J \cdot \nabla H$ for skew-symmetric $J$, the layer performs Hamiltonian flow. If $\Delta\psifield = (M + J)\nabla H$, it performs metriplectic dynamics. The Hamiltonian $H$ encodes \emph{what good features look like}; the dynamics encodes \emph{how to get there}.

\paragraph{Transformer attention as a dissipative system.}
Self-attention computes $\Delta\mathbf{h}_i = \sum_j \softmax(Q_i \cdot K_j / \sqrt{d})_j \cdot V_j$. This is the gradient of a LogSumExp energy:
\begin{equation}
    H_{\text{attn}}(\mathbf{h}) = -\sum_i \log \sum_j \exp\!\bigl(\mathbf{h}_i^\top W_Q W_K^\top \mathbf{h}_j / \sqrt{d}\bigr),
    \label{eq:H_attention}
\end{equation}
whose gradient is precisely the softmax-weighted value aggregation (cf.\ modern Hopfield networks~\citep{ramsauer2021hopfield}). The MLP block adds a second energy $H_{\text{MLP}} = -\sum_i \|\ReLU(W_{\text{up}} \mathbf{h}_i)\|^2$. Both are \emph{purely dissipative}: every layer descends its energy landscape. There is no conservative component---energy is not preserved across layers.

The attention Hamiltonian is \emph{nonlinear} (LogSumExp $\to$ softmax), which gives transformers their expressiveness. The softmax creates competitive dynamics: the most similar token dominates, others are suppressed. This winner-take-all behaviour is the source of attention's power---and it arises from the \emph{shape} of the energy landscape, not from the dynamics.

\paragraph{Metriplector as a designed metriplectic system.}
Metriplector operates in a compressed field space ($K{=}32$) where the complete set of pairwise field interactions is tractable.
During dynamics, $J$ couples all $K$ fields via matrix multiply; at readout, the stress-energy tensor $T^{\mu\nu}$ extracts all $K^2 = 1{,}024$ pairwise features ($K$ diagonal $+$ $K(K{-}1)/2$ symmetric $+$ $K(K{-}1)/2$ antisymmetric).
The symmetric-antisymmetric decomposition is preserved by construction: the metric tensor $M$ (symmetric, PSD) governs diffusion, while the Poisson bracket $J$ (skew-symmetric) governs advection.
These roles are enforced architecturally, not learned.

The key structural differences between attention and Metriplector are summarised in Table~\ref{tab:hamiltonian_comparison}.

\begin{table}[ht]
\centering
\caption{\textbf{Attention vs.\ Metriplector as dynamical systems.} Each layer's update $\Delta\psifield$ can be decomposed into dynamics type (how it moves on the energy landscape) and Hamiltonian type (what the landscape looks like). $D$ = representation dimension, $K$ = field dimension, $N$ = spatial positions.}
\label{tab:hamiltonian_comparison}
\small
\begin{tabular}{lcc}
\toprule
& \textbf{Attention} & \textbf{Metriplector} \\
\midrule
\textbf{Dynamics} & Pure dissipation ($\nabla H$ only) & Designed metriplectic ($M{+}J$) \\[3pt]
\textbf{Hamiltonian} & LogSumExp (nonlinear) & Spatial (quadratic) + stress-energy \\[3pt]
\textbf{Spatial coupling} & Global (all-to-all in standard form) & Local, $O(9N)$ DWConv $3{\times}3$ \\[3pt]
\textbf{Channel mixing} & MLP ($O(D^2)$) & Complete $J$ + stress-energy ($O(K^2)$) \\[3pt]
\textbf{Conservation} & None & Energy conservation by antisymmetric $J$ \\[3pt]
\textbf{Cross-field coverage} & --- & All $K(K{-}1)/2$ pairs \\
\bottomrule
\end{tabular}
\end{table}

The comparison reveals a design space along two axes: the \emph{expressiveness} of the Hamiltonian (how nonlinear is $H$?) and the \emph{structure} of the dynamics (how much of the metriplectic decomposition is preserved?).
Attention has the most expressive $H$ (LogSumExp) with purely dissipative dynamics.
Metriplector preserves the full metriplectic structure while drawing its nonlinearity from the physics itself---the composition of metriplectic evolution steps across layers, modulated by content-dependent operators.

By construction, the composition of metriplectic evolution steps across layers can approximate any port-Hamiltonian system with dissipation while preserving structural laws---the antisymmetric Poisson bracket conserves energy exactly, and the symmetric positive-semidefinite metric tensor guarantees entropy production.
Information propagation on the latent graph is bounded: the screened Laplacian's exponential decay implies that recurrent rounds are necessary and sufficient for long-range reasoning, analogous to the Lieb-Robinson bound in lattice systems~\citep{lieb1972finite,nachtergaele2006lieb}.

\section{The Metriplector Architecture}
\label{sec:method}

\subsection{Dissipative Branch: Screened Poisson on Graphs}

The dissipative branch of the GENERIC equation, shared across spatial reasoning domains, takes the form:
\begin{equation}
    (L_W + \diag(\Lambda_k))\psifield_k = \mathbf{b}_k, \quad k = 1, \ldots, K,
    \label{eq:poisson}
\end{equation}
where $L_W$ is the weighted graph Laplacian with learned edge conductances:
\begin{equation}
    w_{ij} = \softplus(\mathbf{h}_i^\top W_\text{sym} \mathbf{h}_j), \quad W_\text{sym} = \ReLU\!\left(\tfrac{1}{2}(W_\text{raw} + W_\text{raw}^\top)\right).
    \label{eq:conductance}
\end{equation}

\begin{proposition}[Well-posedness]
\label{prop:wellposed}
For $\Lambda_k > 0$ (via softplus) and $w_{ij} > 0$ (via softplus on bilinear form), $A_k = L_W + \diag(\Lambda_k)$ is symmetric positive definite, guaranteeing a unique solution.
\end{proposition}

The system is solved via CG~\citep{hestenes1952methods} with 40--60 iterations.
A custom \texttt{torch.autograd.Function} implements implicit differentiation via the adjoint method, requiring only $O(N)$ memory.
An optimized Rust kernel~\citep{matsakis2014rust} with Rayon parallelism provides $\sim$7$\times$ speedup.

\subsection{Recurrent Multigrid Architecture}
\label{sec:multigrid}

The full architecture (Figure~\ref{fig:architecture}) repeats a two-level multigrid V-cycle for $R$ rounds:

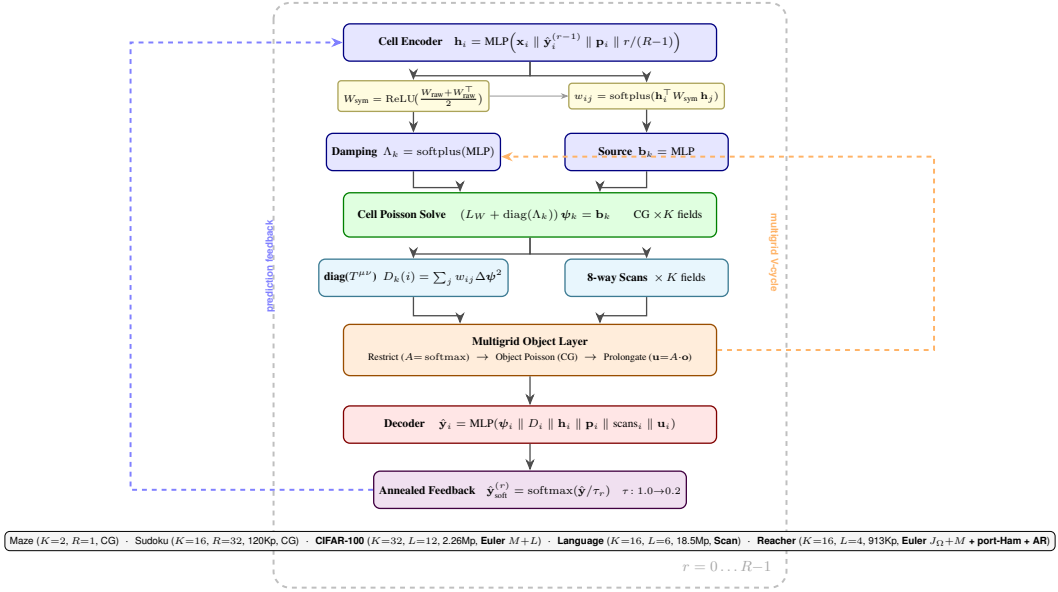
\begin{figure}[t]
\centering
\resizebox{\textwidth}{!}{%
\begin{tikzpicture}[
    block/.style={draw, thick, rounded corners=4pt, minimum height=0.8cm, minimum width=2.8cm, align=center, font=\scriptsize},
    phys/.style={block, fill=green!12, draw=green!50!black},
    enc/.style={block, fill=blue!10, draw=blue!50!black},
    feat/.style={block, fill=cyan!8, draw=cyan!50!black},
    obj/.style={block, fill=orange!14, draw=orange!60!black},
    dec/.style={block, fill=red!10, draw=red!50!black},
    fb/.style={block, fill=violet!12, draw=violet!50!black},
    param/.style={draw, thick, rounded corners=2pt, minimum height=0.55cm, align=center, fill=yellow!15, draw=yellow!60!black, font=\tiny},
    arr/.style={-{Stealth[length=2.5mm, width=2mm]}, thick, black!70},
    feedback/.style={-{Stealth[length=2.5mm, width=2mm]}, very thick, dashed},
]
\draw[rounded corners=10pt, very thick, black!25, dashed] (-5.5, 0.35) rectangle (5.5, -12.2);
\node[font=\footnotesize\bfseries\sffamily, black!35, anchor=south east] at (5.3, -12.0) {$r = 0 \ldots R{-}1$};

\node[enc, minimum width=8cm] (cellenc) at (0, -0.5) {%
\textbf{Cell Encoder}\quad $\mathbf{h}_i = \text{MLP}\!\left(\mathbf{x}_i \;\|\; \hat{\mathbf{y}}^{(r-1)}_i \;\|\; \mathbf{p}_i \;\|\; r/(R{-}1)\right)$};

\node[param, minimum width=3.2cm] (wsym) at (-2.5, -1.65) {$W_\text{sym} = \ReLU\!\bigl(\tfrac{W_\text{raw}+W_\text{raw}^\top}{2}\bigr)$};
\node[param, minimum width=3.2cm] (wedge) at (2.5, -1.65) {$w_{ij} = \softplus(\mathbf{h}_i^\top W_\text{sym}\, \mathbf{h}_j)$};
\draw[arr] (cellenc.south) -- ++(0,-0.3) -| (wsym.north);
\draw[arr] (cellenc.south) -- ++(0,-0.3) -| (wedge.north);
\draw[-{Stealth[length=1.5mm]}, thin, black!40] (wsym.east) -- (wedge.west);

\node[enc, minimum width=3.5cm] (damp) at (-2.5, -2.85) {\textbf{Damping}\; $\Lambda_k = \softplus(\text{MLP})$};
\node[enc, minimum width=3.5cm] (source) at (2.5, -2.85) {\textbf{Source}\; $\mathbf{b}_k = \text{MLP}$};

\draw[arr] (wsym.south) -- ++(0,-0.35) -| (damp.north);
\draw[arr] (wedge.south) -- ++(0,-0.35) -| (source.north);

\node[phys, minimum width=8cm, minimum height=0.95cm] (poisson) at (0, -4.2) {%
\textbf{Cell Poisson Solve} \quad $(L_W + \diag(\Lambda_k))\,\psifield_k = \mathbf{b}_k$%
\qquad CG $\times K$ fields};

\draw[arr] (damp.south) -- ++(0,-0.35) -| ([xshift=-1.5cm]poisson.north);
\draw[arr] (source.south) -- ++(0,-0.35) -| ([xshift=1.5cm]poisson.north);

\node[feat, minimum width=3.5cm] (dissip) at (-2.5, -5.55) {\textbf{diag($T^{\mu\nu}$)}\; $D_k(i) = \sum_j w_{ij}\Delta\psifield^2$};
\node[feat, minimum width=3.5cm] (scans) at (2.5, -5.55) {\textbf{8-way Scans}\; $\times\, K$ fields};

\draw[arr] (poisson.south) -- ++(0,-0.35) -| (dissip.north);
\draw[arr] (poisson.south) -- ++(0,-0.35) -| (scans.north);

\node[obj, minimum width=8cm, minimum height=1.1cm] (objlayer) at (0, -7.1) {%
\begin{tabular}{c}
\textbf{Multigrid Object Layer} \\[1pt]
{\tiny Restrict ($A{=}\softmax$)} $\;\to\;$
{\tiny Object Poisson (CG)} $\;\to\;$
{\tiny Prolongate ($\mathbf{u}{=}A{\cdot}\mathbf{o}$)}
\end{tabular}};

\draw[arr] (dissip.south) -- ++(0,-0.4) -| ([xshift=-1.5cm]objlayer.north);
\draw[arr] (scans.south) -- ++(0,-0.4) -| ([xshift=1.5cm]objlayer.north);

\node[dec, minimum width=8cm] (decoder) at (0, -8.7) {%
\textbf{Decoder} \quad $\hat{\mathbf{y}}_i = \text{MLP}\!\left(\psifield_i \;\|\; D_i \;\|\; \mathbf{h}_i \;\|\; \mathbf{p}_i \;\|\; \text{scans}_i \;\|\; \mathbf{u}_i\right)$};

\draw[arr] (objlayer.south) -- (decoder.north);

\node[fb, minimum width=6cm] (annealed) at (0, -10.1) {%
\textbf{Annealed Feedback}\quad $\hat{\mathbf{y}}_\text{soft}^{(r)} = \softmax\!\left(\hat{\mathbf{y}} / \tau_r\right)$\quad {\tiny$\tau\!: 1.0 {\to} 0.2$}};

\draw[arr] (decoder.south) -- (annealed.north);

\node[draw, rounded corners=3pt, fill=black!4, font=\tiny\sffamily, inner sep=3pt] at (0, -11.2) {%
Maze ($K{=}2$, $R{=}1$, CG) $\;\cdot\;$
Sudoku ($K{=}16$, $R{=}32$, 120Kp, CG) $\;\cdot\;$
\textbf{CIFAR-100} ($K{=}32$, $L{=}12$, 2.26Mp, \textbf{Euler $M{+}L$}) $\;\cdot\;$
\textbf{Language} ($K{=}16$, $L{=}6$, 18.5Mp, \textbf{Scan}) $\;\cdot\;$
\textbf{Reacher} ($K{=}16$, $L{=}4$, 913Kp, \textbf{Euler $J_\Omega{+}M$ + port-Ham + AR})};

\draw[feedback, blue!50] (annealed.west) -- ++(-5.2,0) |- (cellenc.west);
\node[font=\tiny\sffamily\bfseries, blue!50, rotate=90, anchor=south] at (-5.35, -5.3) {prediction feedback};

\draw[feedback, orange!60] (objlayer.east) -- ++(4.65,0) |- ([yshift=-0.1cm]damp.east);
\node[font=\tiny\sffamily\bfseries, orange!60, rotate=-90, anchor=south] at (5.0, -5.0) {multigrid V-cycle};

\end{tikzpicture}%
}
\caption{\textbf{Metriplector architecture.} A single round of the recurrent V-cycle (repeated $R$ times with shared weights). The cell encoder produces per-cell features $\mathbf{h}$ from input, previous predictions, position, and round fraction. Learned symmetric conductances $w_{ij}$ define the graph Laplacian. Damping and source MLPs produce per-cell screening and forcing terms. $K$ independent screened Poisson equations are solved via CG. The dissipation readout $D_k = \sum_j w_{ij}(\psifield_i - \psifield_j)^2 \approx |\nabla\psifield_k|^2$ extracts the diagonal of the stress-energy tensor (the same $T^{\mu\nu}$ readout used in CIFAR-100, cf.\ Section~\ref{sec:discussion}); 8-way directional scans add global context. The multigrid object layer restricts to learned soft-assignment groups, solves a coarse Poisson, and prolongates back. The decoder reads all features to produce per-cell predictions, fed back via annealed softmax ($\tau\!: 1.0 \to 0.2$). \textcolor{blue!50}{Blue}: prediction feedback across rounds. \textcolor{orange!60}{Orange}: multigrid V-cycle within each round. Reasoning domains (maze, Sudoku) share this CG-based architecture with different $K$, $R$, and output dimension $C$; CIFAR-100 uses a distinct Euler-based instantiation (Section~\ref{sec:cifar100_method}); Reacher extends the CIFAR instantiation with port-Hamiltonian action conditioning for world modeling (Section~\ref{sec:reacher_method}).}
\label{fig:architecture}
\end{figure}

\paragraph{Cell encoder.}
$\mathbf{h}_i = \text{MLP}(\mathbf{x}_i \| \hat{\mathbf{y}}^{(r-1)}_i \| \mathbf{p}_i \| r/(R{-}1))$ produces per-cell features incorporating input, previous predictions, position, and round fraction.

\paragraph{Damping and source.}
Both are produced by 3-layer MLPs reading a rich feature vector: $\mathbf{h}$, position, previous $\psifield$, 8-way directional scans, cross-field statistics, and unpooled object features from the previous round's V-cycle.

\paragraph{Directional scans.}
Cumulative sums of normalized $\psifield$ along 8 directions (4 cardinal + 4 diagonal) provide each cell with aggregated row, column, and diagonal information without explicit long-range edges.

\paragraph{Object layer (multigrid V-cycle).}
Cells are softly assigned to $K_\text{obj}$ learned clusters via $A = \softmax(\text{MLP}(\tilde{\psifield} \| \mathbf{p}) / \tau)$.
Cell features are pooled ($A^\top \cdot \mathbf{f}$), a coarse Poisson equation is solved on the object graph, and results are prolongated back ($A \cdot \mathbf{o}^\text{out}$).
This implements a classical two-level V-cycle~\citep{briggs2000multigrid} with \emph{learned} restriction and prolongation operators.

\paragraph{Decoder.}
An MLP maps the full feature set (fields, dissipation, cell features, position, scans, unpooled objects) to task-specific outputs.

\paragraph{Prediction feedback.}
Softmax predictions are fed back with linearly annealing temperature: $\tau_r = 1.0 \cdot (1 - r/(R{-}1)) + 0.2 \cdot r/(R{-}1)$.
Early rounds explore; late rounds commit---analogous to simulated annealing~\citep{kirkpatrick1983optimization}.

\subsection{Domain-Specific Adaptations}

\begin{table}[ht]
\centering
\caption{\textbf{Domain configurations.} All variants share the metriplectic formalism; domains differ in which GENERIC channels are active, the solver, and domain-specific adaptations. Sudoku and CIFAR-100 implement the same underlying physics (Section~\ref{sec:discussion}): both read $T^{\mu\nu}$ from evolved fields. Reacher extends this to action-conditioned dynamics.}
\label{tab:domains}
\small
\begin{tabular}{lccccccl}
\toprule
\textbf{Domain} & \textbf{Graph} & $K$ & \textbf{Depth} & $K_\text{obj}$ & \textbf{Output} & \textbf{Params} & \textbf{Dynamics} \\
\midrule
\textbf{CIFAR-100} & 2D grid 16$\times$16 & 32 & $L{=}12$ & --- & 100 classes & 2.26M & \textbf{Euler, $M{+}L$} \\
\textbf{Reacher} & 2D grid 8$\times$8 & 16 & $L{=}4$ & --- & 8$\times$8$\times$128 latent & 913K & \textbf{Euler, $J_\Omega{+}M$, port-Ham, AR} \\
Sudoku & 8-conn lattice & 16 & $R{=}32$ & 16 & 9 digits & 120K & CG, $M$ only \\
\textbf{Language} & Causal 1D chain & 16 & $L{=}6$ & --- & 1024 tokens & 18.5M & \textbf{Scan, $M$ causal} \\
Maze & 4-conn grid & 2 & $R{=}1$ & --- & 5 classes & 43.8K & CG, $M$ only \\
\bottomrule
\end{tabular}
\end{table}

\paragraph{Maze.}
The simplest variant: a single Poisson solve ($R=1$).
Cell types are encoded as anonymous integer indices---the model never receives labels like ``wall'' or ``corridor.''
The conductance matrix $W \in \R^{4 \times 4}$ discovers wall/corridor structure from data alone.
We introduce \emph{$\lambda/N$ scaling}---normalizing the damping term by the number of nodes---to keep the spectral balance of the graph Laplacian constant across grid sizes.

\paragraph{Sudoku.}
The richest variant: 32 recurrent multigrid rounds with $K=16$ fields and $K_\text{obj}=16$ objects.
Digits are encoded as anonymous indices; the graph is a pure 8-connected spatial lattice with no row/column/box edges.
All Sudoku structure must be discovered by the model from data alone (zero structural injection---no row, column, or box edges are provided; the model receives digit values but no encoding of the constraint rules).
The feature readout $D_k(i) = \sum_j w_{ij}(\psifield_k(i) - \psifield_k(j))^2$ is the discrete analog of the gradient energy $|\nabla\psifield_k|^2$---equivalently, the diagonal of the stress-energy tensor $T_{kk}^{\mu\mu}$ that the CIFAR-100 variant computes via learned gradient convolutions (Section~\ref{sec:discussion}).
The object layer's soft assignment $\rho = \softmax(\text{MLP}(\psifield, \mathbf{p})/\tau)$ with coarse-graining $\psifield^{(1)} = \rho^\top\psifield^{(0)}$ performs learned spatial pooling analogous to block-spin coarse-graining.
Difficulty-stratified sampling overweights hard puzzles during training.

\subsection{CIFAR-100: Full Metriplectic Layers}
\label{sec:cifar100_method}

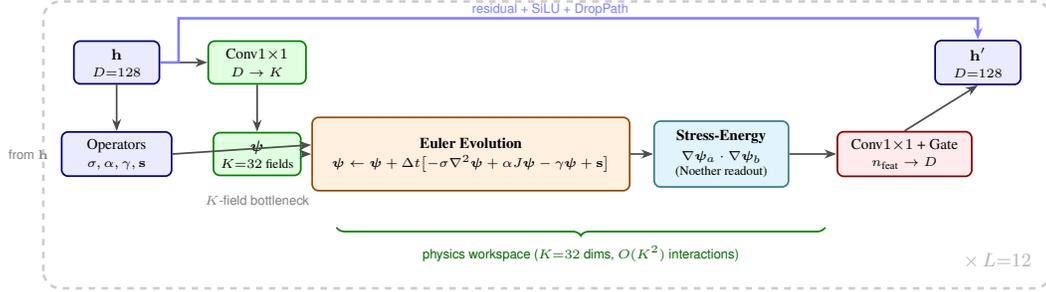
\begin{figure}[t]
\centering
\resizebox{\textwidth}{!}{%
\begin{tikzpicture}[
    hblock/.style={draw, thick, rounded corners=3pt, minimum height=0.7cm, fill=blue!8, draw=blue!50!black, align=center, font=\scriptsize},
    psiblock/.style={draw, thick, rounded corners=3pt, minimum height=0.7cm, fill=green!12, draw=green!50!black, align=center, font=\scriptsize},
    physblock/.style={draw, thick, rounded corners=3pt, minimum height=0.9cm, fill=orange!12, draw=orange!60!black, align=center, font=\scriptsize},
    readblock/.style={draw, thick, rounded corners=3pt, minimum height=0.7cm, fill=cyan!10, draw=cyan!50!black, align=center, font=\scriptsize},
    gateblock/.style={draw, thick, rounded corners=3pt, minimum height=0.7cm, fill=red!8, draw=red!50!black, align=center, font=\scriptsize},
    arr/.style={-{Stealth[length=2mm, width=1.5mm]}, thick, black!70},
    dimlabel/.style={font=\tiny\sffamily, black!50},
]

\draw[rounded corners=8pt, very thick, black!20, dashed] (-0.5, -3.2) rectangle (16.0, 1.5);
\node[font=\footnotesize\bfseries\sffamily, black!30, anchor=south east] at (15.8, -3.0) {$\times\, L{=}12$};

\node[hblock, minimum width=1.4cm] (hin) at (0.7, 0.5) {$\mathbf{h}$\\{\tiny$D{=}128$}};

\node[psiblock, minimum width=1.6cm] (h2psi) at (3.0, 0.5) {Conv$1{\times}1$\\{\tiny$D \to K$}};
\draw[arr] (hin.east) -- (h2psi.west);

\node[psiblock, minimum width=1.4cm] (psi) at (3.0, -1.0) {$\psifield$\\{\tiny$K{=}32$ fields}};
\draw[arr] (h2psi.south) -- (psi.north);

\node[hblock, minimum width=1.8cm] (ops) at (0.7, -1.0) {Operators\\{\tiny$\sigma, \alpha, \gamma, \mathbf{s}$}};
\draw[arr] (hin.south) -- (ops.north);
\node[dimlabel, anchor=east] at (-0.3, -1.0) {from $\mathbf{h}$};

\node[physblock, minimum width=4.0cm, minimum height=1.2cm] (euler) at (6.5, -1.0) {%
\begin{tabular}{c}
\textbf{Euler Evolution}\\[1pt]
{\tiny $\psifield \leftarrow \psifield + \Delta t\bigl[-\sigma\nabla^2\psifield + \alpha J\psifield - \gamma\psifield + \mathbf{s}\bigr]$}
\end{tabular}};
\draw[arr] (psi.east) -- (euler.west);
\draw[arr] (ops.east) -- ([yshift=0.15cm]euler.west);

\node[readblock, minimum width=2.2cm, minimum height=1.0cm] (readout) at (10.6, -1.0) {%
\begin{tabular}{c}
\textbf{Stress-Energy}\\[1pt]
{\tiny $\nabla\psifield_a \cdot \nabla\psifield_b$}\\[-1pt]
{\tiny (Noether readout)}
\end{tabular}};
\draw[arr] (euler.east) -- (readout.west);

\node[gateblock, minimum width=2.2cm] (proj) at (13.6, -1.0) {Conv$1{\times}1$ + Gate\\{\tiny$n_\text{feat} \to D$}};
\draw[arr] (readout.east) -- (proj.west);

\node[hblock, minimum width=1.4cm] (hout) at (14.8, 0.5) {$\mathbf{h}'$\\{\tiny$D{=}128$}};
\draw[arr] (proj.north) -- (hout.south);

\draw[arr, blue!50, very thick] (hin.east) -- ++(0.3,0) |- ([yshift=0.35cm]hout.north) -- (hout.north);
\node[dimlabel, blue!50, anchor=south] at (7.8, 1.15) {residual + SiLU + DropPath};

\node[dimlabel] at (3.0, -1.75) {$K$-field bottleneck};

\draw[decorate, decoration={brace, amplitude=4pt, mirror}, thick, green!50!black]
    (4.3, -2.2) -- (12.3, -2.2) node[midway, below=5pt, font=\tiny\sffamily, green!40!black] {physics workspace ($K{=}32$ dims, $O(K^2)$ interactions)};

\end{tikzpicture}%
}
\caption{\textbf{CIFAR-100 metriplectic layer} ($\times 12$, non-shared weights). The representation $\mathbf{h}$ ($D{=}128$) flows along the top via residual connections. Each layer projects $\mathbf{h}$ down to $K{=}32$ physics fields $\psifield$, evolves them under the full metriplectic equation (diffusion $+$ advection $+$ damping $+$ source), extracts physically meaningful features via the stress-energy tensor (Noether readout), and projects back to $D$ via gated mixing. The $K$-field bottleneck reduces pairwise interaction cost by $16\times$ compared to operating in the full $D$ space.}
\label{fig:cifar100_arch}
\end{figure}

Image classification requires a fundamentally different instantiation of the metriplectic formalism (Figure~\ref{fig:cifar100_arch}).
Rather than solving for equilibrium (maze, Sudoku), the CIFAR-100 architecture evolves $K$ physics fields $\psifield$ through 12 \emph{non-shared} layers of Euler integration, with a separate representation $\mathbf{h} \in \R^D$ ($D{=}128$) carrying class information via residual connections.

\paragraph{The operator-from-input principle.}
The representation $\mathbf{h}$ defines the physics that $\psifield$ evolves under---there is no circular dependence.
Five Conv$1{\times}1$ projections derive operator coefficients from $\mathbf{h}$:
\begin{align}
    \psifield &= W_\psi \, \mathbf{h}, &
    \sigma &= \softplus(W_\sigma \, \mathbf{h}), \nonumber \\
    \alpha &= W_\alpha \, \mathbf{h}, &
    \gamma &= \softplus(W_\gamma \, \mathbf{h}) + 0.1, \nonumber \\
    \mathbf{s} &= \text{clamp}(W_s \, \mathbf{h}, \pm 5), & &
    \label{eq:operator_from_h}
\end{align}
where each $W_\cdot$ is a learned $K \times D$ matrix. The field $\psifield$ is created fresh from $\mathbf{h}$ in every layer. Ablations confirm this design: recurrent $\psifield$ degrades accuracy by 3.4 points, and letting $\psifield$ build its own operator costs 14.3 points (Table~\ref{tab:cifar100_ablations}).

\paragraph{The $K$-field bottleneck.}
The representation $\mathbf{h} \in \R^D$ ($D{=}128$) carries class information via residual connections; the physics fields $\psifield \in \R^K$ ($K{=}32$) are where metriplectic dynamics operate.
This mirrors multi-head attention: project to a small space, compute complete interactions ($K^2 = 1{,}024$ stress-energy features), and project back.
The Conv$1{\times}1$ projection back to $D$ selects which physics features matter for classification.

\paragraph{Readout via Noether's theorem.}
After metriplectic evolution, the model must extract features from the evolved fields $\psifield$ and project them back into the representation $\mathbf{h}$. Rather than treating this as an arbitrary feature-engineering step, we derive the readout from \emph{Noether's theorem}: every continuous symmetry of the dynamics yields a conserved current, and these conserved currents are the natural observables of the field theory.

\textbf{Spatial translation symmetry} yields the \emph{stress-energy tensor}:
\begin{equation}
    T_{ab}^{ij} = \partial_i \psifield_a \cdot \partial_j \psifield_b,
    \label{eq:stress_tensor}
\end{equation}
where $a, b$ index fields and $i, j$ index spatial directions. We decompose this into physically meaningful components:
\begin{itemize}
    \item \textbf{Energy density} (symmetric trace): $E_{ab} = \nabla\psifield_a \cdot \nabla\psifield_b$. For $a{=}b$: edge strength of field $a$ (gradient magnitude). For $a{\neq}b$: whether fields $a$ and $b$ have co-located edges---a structural correlation measure.
    \item \textbf{Vorticity} (antisymmetric shear): $V_{ab} = \partial_x \psifield_a \cdot \partial_y \psifield_b - \partial_x \psifield_b \cdot \partial_y \psifield_a$. Non-zero at corners and junctions, where the gradient of one field is perpendicular to another's. Zero along straight edges. Detects structural complexity.
\end{itemize}
The directional gradients $\partial_x \psifield$, $\partial_y \psifield$ are computed by learned depthwise convolutions (initialized as Sobel-like filters). The total feature count is $K^2$: $K$ diagonal energy terms + $K(K{-}1)/2$ cross-field energy terms + $K(K{-}1)/2$ vorticity terms.

We also evaluate two alternative readouts:
\begin{itemize}
    \item \textbf{Field-curvature products}: $\psifield_a \cdot \Delta\psifield_b$ (symmetric, field value $\times$ Laplacian) and $\psifield_a \Delta\psifield_b - \psifield_b \Delta\psifield_a$ (antisymmetric, cross-field twist), yielding $K + K(K{-}1)/2$ features. These mix field \emph{values} with \emph{second} derivatives---related to the stress tensor through integration by parts but not identical. The stress tensor uses \emph{first} derivative products, which are sign-consistent (always non-negative for $a{=}b$) and geometrically cleaner.
    \item \textbf{Noether currents} (full symmetry set): momentum density $\psifield \cdot \nabla\psifield$ (translation), angular momentum $\mathbf{x} \times \mathbf{p}$ (rotation), dilation current $\mathbf{x} \cdot \mathbf{p}$ (scale), plus stress-energy. These incorporate spatial coordinates $(\mathbf{x}, \mathbf{y})$, providing position-aware features---a physics-native positional encoding. Yields $5K + 3K(K{-}1)/2$ features.
\end{itemize}

The stress-energy readout achieves the best accuracy (Table~\ref{tab:cifar100}), confirming that the conserved quantities of the spatial symmetry group are the most informative observables of the evolved field.

A Conv$1{\times}1$ projection ($n_\text{feat} \to D$) with BatchNorm~\citep{ioffe2015batch} and learned gating mixes the physics features back into $\mathbf{h}$ via a residual connection with SiLU activation~\citep{elfwing2018sigmoid} and stochastic depth~\citep{huang2016deep}.

\subsection{World Modeling: Metriplectic Dynamics for Robotic Control}
\label{sec:reacher_method}

The Reacher domain introduces a fundamentally new challenge for the metriplectic formalism: \emph{action-conditioned latent dynamics}.
Rather than classifying a static input (CIFAR-100) or satisfying constraints (Sudoku), the world model must predict how the environment state evolves in response to control actions---a task that requires the architecture to model \emph{controllable} physical systems.

\begin{figure}[t]
\centering
\resizebox{\textwidth}{!}{%
\begin{tikzpicture}[
    block/.style={draw, thick, rounded corners=3pt, minimum height=0.8cm, align=center, font=\scriptsize},
    encblock/.style={block, fill=purple!10, draw=purple!50!black},
    hblock/.style={block, fill=blue!8, draw=blue!50!black},
    psiblock/.style={block, fill=green!12, draw=green!50!black},
    physblock/.style={block, fill=orange!12, draw=orange!60!black},
    actblock/.style={block, fill=red!8, draw=red!50!black},
    readblock/.style={block, fill=cyan!10, draw=cyan!50!black},
    arr/.style={-{Stealth[length=2mm, width=1.5mm]}, thick, black!70},
    dimlabel/.style={font=\tiny\sffamily, black!50},
]

\node[encblock, minimum width=2.0cm] (img) at (0, 0) {Image $\mathbf{I}_t$\\{\tiny$64{\times}64{\times}3$}};
\node[encblock, minimum width=2.2cm] (enc) at (3.0, 0) {ConvEncoder\\{\tiny 6 layers, $\sim$560K}};
\draw[arr] (img.east) -- (enc.west);

\node[hblock, minimum width=1.8cm] (h) at (6.0, 0) {$\mathbf{h}_t$\\{\tiny$8{\times}8{\times}128$}};
\draw[arr] (enc.east) -- (h.west);

\node[actblock, minimum width=1.8cm] (act) at (3.0, -2.0) {Action $\mathbf{a}_t$\\{\tiny$\R^{10}$}};
\node[actblock, minimum width=2.0cm] (actmlp) at (6.0, -2.0) {MLP\\{\tiny$10 \to d_\text{act}$}};
\draw[arr] (act.east) -- (actmlp.west);

\draw[rounded corners=6pt, very thick, black!20, dashed] (8.0, -3.0) rectangle (15.5, 1.2);
\node[font=\footnotesize\bfseries\sffamily, black!30, anchor=south east] at (15.3, -2.8) {$\times\, L{=}4$};

\node[psiblock, minimum width=1.6cm] (psi) at (9.2, 0) {$\psifield = W_\psi \mathbf{h}$\\{\tiny$K{=}16$, 8$(q,p)$ pairs}};
\draw[arr] (h.east) -- (psi.west);

\node[actblock, minimum width=2.0cm] (opmod) at (9.2, -2.0) {$\delta\sigma, \delta\alpha,$\\$\delta\gamma, \delta\mathbf{s}$\\{\tiny port-Ham}};
\draw[arr] (actmlp.east) -- (opmod.west);

\node[physblock, minimum width=2.8cm, minimum height=1.2cm] (euler) at (12.0, -0.5) {%
\begin{tabular}{c}
\textbf{MetriplectorLayer}\\[1pt]
{\tiny $\psifield + \Delta t[-\sigma\nabla^2\psifield$}\\[-2pt]
{\tiny $+ \alpha J_\Omega\psifield - \gamma\psifield + \mathbf{s}]$}
\end{tabular}};
\draw[arr] (psi.east) -- ([yshift=0.3cm]euler.west);
\draw[arr] (opmod.east) -- ([yshift=-0.3cm]euler.west);

\node[readblock, minimum width=1.6cm] (readout) at (14.8, 0) {$T^{\mu\nu}$\\{\tiny Noether}};
\draw[arr] ([yshift=0.3cm]euler.east) -- (readout.west);

\node[hblock, minimum width=1.8cm] (hout) at (17.0, 0) {$\hat{\mathbf{h}}_{t+1}$\\{\tiny$8{\times}8{\times}128$}};
\draw[arr] (readout.east) -- (hout.west);

\draw[arr, blue!50, very thick] (h.north) -- ++(0,0.7) -| (hout.north);
\node[dimlabel, blue!50] at (11.5, 1.0) {residual + GroupNorm};

\node[encblock, minimum width=2.2cm, dashed] (ema) at (17.0, -2.0) {EMA Target\\{\tiny$f_\text{enc}^\text{target}(\mathbf{I}_{t+1})$}};
\draw[arr, dashed, black!40] (ema.north) -- node[right, font=\tiny, black!50] {$\Lcal_\text{MSE}$} (hout.south);

\end{tikzpicture}%
}
\caption{\textbf{Reacher world model architecture.} A convolutional encoder maps $64{\times}64$ images to an $8{\times}8{\times}128$ latent grid. The action modulates PDE operators via port-Hamiltonian conditioning (zero-initialized). Four MetriplectorLayers evolve fresh $\psifield$ fields---organized as 8 conjugate $(q,p)$ pairs with canonical symplectic $J_\Omega$---and extract $T^{\mu\nu}$ features. Multi-step AR training (scheduled warmup) closes the exposure bias gap. The prediction target comes from an EMA target encoder. At evaluation, CEM plans by rolling out action sequences through the latent dynamics.}
\label{fig:reacher_arch}
\end{figure}
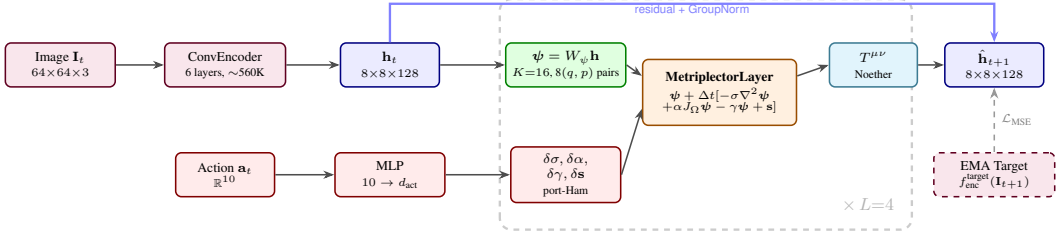

This instantiation follows the CIFAR-100 pattern (stacked layers, fresh $\psifield$ per layer, full metriplectic dynamics) because world models require hierarchical feature extraction, not equilibrium-seeking convergence.
The key extensions beyond CIFAR-100 are a high-capacity learned encoder, port-Hamiltonian action conditioning, a canonical symplectic Poisson bracket that pairs the $K$ fields into $(q,p)$ conjugate pairs, and multi-step autoregressive training with scheduled warmup for stable long-horizon rollouts (Figure~\ref{fig:reacher_arch}).

\paragraph{Encoder: image to latent grid.}
A 6-layer convolutional encoder compresses $64{\times}64{\times}3$ RGB images into an $8{\times}8{\times}128$ latent grid $\mathbf{h}$:
\begin{equation}
    \mathbf{h} = f_\text{enc}(\mathbf{I}): \R^{64 \times 64 \times 3} \to \R^{8 \times 8 \times 128},
    \label{eq:reacher_encoder}
\end{equation}
using three stride-2 stages, each with two convolutions, GroupNorm, and SiLU activations, progressively widening through $64 \to 128 \to D$ channels ($\sim$560K parameters).
Two copies are maintained: an \emph{online encoder} trained via backpropagation, and an \emph{EMA target encoder} ($\tau{=}0.996$) providing prediction targets.
Self-supervised latent prediction is stabilised by variance regularization ($\lambda_\text{var}{=}0.50$, penalising per-channel variance below 1.0) and covariance regularization ($\lambda_\text{cov}{=}0.04$, decorrelating channels).

\paragraph{Port-Hamiltonian action conditioning.}
The action $\mathbf{a} \in \R^{10}$ (5 frameskip $\times$ 2D joint torques) modulates the PDE operators rather than the state, following port-Hamiltonian control theory~\citep{desai2021port}:
\begin{equation}
    \mathbf{e}_a = \text{MLP}(\mathbf{a}): \R^{10} \to \R^{d_\text{act}}, \qquad
    \delta\sigma, \delta\alpha, \delta\gamma, \delta\mathbf{s} = W_\sigma \mathbf{e}_a,\; W_\alpha \mathbf{e}_a,\; W_\gamma \mathbf{e}_a,\; W_s \mathbf{e}_a,
    \label{eq:portham_action}
\end{equation}
where each $W_\cdot$ is zero-initialized so the model starts as if no action is applied.
The action modifies the PDE coefficients---diffusion $\sigma$, advection $\alpha$, damping $\gamma$, and source $\mathbf{s}$---rather than the latent state $\mathbf{h}$ directly.
This is the $G \cdot \mathbf{u}$ term from port-Hamiltonian theory: the action configures the physical operators, and the dynamics computes the response.
Because $\mathbf{h}$ passes through unchanged, multi-step autoregressive rollouts remain norm-stable without additional tuning.

\paragraph{Symplectic MetriplectorLayer.}
Each of $L{=}4$ non-shared layers creates fresh fields $\psifield = W_\psi \mathbf{h}$ with $K{=}16$ fields organized as 8 conjugate $(q_i, p_i)$ pairs, evolves them through 2 substeps of the metriplectic equation, and reads out the stress-energy tensor $T^{\mu\nu}$.
The Poisson bracket uses a \emph{canonical symplectic} structure: a fixed block-diagonal matrix $J_\Omega = \text{blockdiag}\bigl(\begin{smallmatrix}0&+1\\-1&0\end{smallmatrix}\bigr)$ enforces Hamiltonian coupling between each $(q_i, p_i)$ pair, supplemented by a learnable antisymmetric cross-pair coupling $J_\text{cross} = W - W^\top$ (initialized small). The full Poisson bracket $J = J_\Omega + J_\text{cross}$ ensures that the advective dynamics $\alpha J \psifield$ instantiates Hamilton's equations on the field manifold while allowing the network to learn inter-pair interactions.
The advection strength $\alpha$ is projected to a \emph{scalar} per spatial position, enforcing uniform Hamiltonian coupling magnitude across all pairs.
A step input norm (GroupNorm at the top of each dynamics step) and per-layer GroupNorm after each residual connection ensure input-scale-invariance and stable latent norms across all CEM rollout steps.

\paragraph{Multi-step autoregressive training.}
Pure teacher forcing---where each step receives the ground-truth previous state---creates an \emph{exposure bias}: during CEM evaluation the model rolls out from its own predictions, encountering the accumulated errors it never saw in training.
We address this with scheduled multi-step autoregressive (AR) training.
During a warmup phase (epochs~1--10), training uses pure teacher forcing over sequences of length~6:
\begin{equation}
    \hat{\mathbf{z}}_t = f_\text{step}(f_\text{enc}^\text{target}(\mathbf{I}_{t-1}), \mathbf{a}_{t-1}), \quad \Lcal_\text{TF} = \sum_{t=1}^{5} \|\hat{\mathbf{z}}_t - f_\text{enc}^\text{target}(\mathbf{I}_t)\|^2.
\end{equation}
After warmup, the first $S{=}3$ steps switch to autoregressive mode: $\hat{\mathbf{z}}_1$ is computed from the encoded observation, then $\hat{\mathbf{z}}_2 = f_\text{step}(\hat{\mathbf{z}}_1, \mathbf{a}_1)$ feeds the model's own prediction back as input, and so on.
Any remaining steps ($t > S$) revert to teacher forcing.
This hybrid schedule lets the model learn to recover from its own prediction errors while maintaining stable gradients via the teacher-forced tail.
At epoch~10 (first AR epoch), step-5 MSE drops from 0.432 to 0.108---a 4$\times$ improvement in multi-step prediction stability.

\paragraph{Evaluation.}
At evaluation, the Cross-Entropy Method~\citep{rubinstein1999cross} plans in latent space: 300 candidate action sequences are sampled, rolled out through the model's dynamics, scored by distance to the goal latent, and iteratively refined over 30 iterations with 30 elite samples.
Success is measured by \texttt{qpos\_match}: all joint angles within 0.05 radians of the target, matching the LeWM protocol exactly~\citep{maes2024lewm}.

\subsection{Language Modeling: Causal Metriplectic Dynamics}
\label{sec:lm_method}

\begin{figure}[t]
\centering
\resizebox{\textwidth}{!}{%
\begin{tikzpicture}[
    block/.style={draw, thick, rounded corners=3pt, minimum height=0.7cm, align=center, font=\scriptsize},
    emb/.style={block, fill=blue!8, draw=blue!50!black},
    scan/.style={block, fill=green!12, draw=green!50!black},
    mg/.style={block, fill=orange!12, draw=orange!60!black},
    mlp/.style={block, fill=red!8, draw=red!50!black},
    arr/.style={-{Stealth[length=2mm, width=1.5mm]}, thick, black!70},
    causal/.style={-{Stealth[length=1.5mm, width=1.2mm]}, thick, green!60!black},
    dimlabel/.style={font=\tiny\sffamily, black!50},
]

\draw[rounded corners=8pt, very thick, black!20, dashed] (2.8, -2.6) rectangle (13.2, 1.8);
\node[font=\footnotesize\bfseries\sffamily, black!30, anchor=south east] at (13.0, -2.4) {$\times\, L{=}6$};

\node[emb, minimum width=2.2cm] (embed) at (1.0, 0.5) {Token Embed\\{\tiny + trigram hash}};

\node[scan, minimum width=2.6cm, minimum height=1.0cm] (scanblk) at (4.8, 0.5) {%
\begin{tabular}{c}
\textbf{Causal Scan}\\[1pt]
{\tiny $\psifield_i = \alpha_i \psifield_{i-1} + \beta_i$}\\[-1pt]
{\tiny $O(N \log N)$ parallel}
\end{tabular}};
\draw[arr] (embed.east) -- (scanblk.west);

\foreach \x in {0,1,2,3,4} {
    \fill[green!40!black] (3.6+\x*0.5, -0.6) circle (2pt);
}
\foreach \x in {0,1,2,3} {
    \draw[causal] (3.7+\x*0.5, -0.6) -- (4.0+\x*0.5, -0.6);
}
\node[dimlabel] at (5.1, -0.95) {strict causality};

\node[emb, minimum width=1.8cm] (srcdamp) at (4.8, -1.7) {Source / Damping\\{\tiny from $\mathbf{h}$, $\psifield_\text{prev}$}};
\draw[arr] (srcdamp.north) -- ([yshift=-0.15cm]scanblk.south);

\node[mg, minimum width=2.8cm, minimum height=1.0cm] (mgrid) at (8.2, 0.5) {%
\begin{tabular}{c}
\textbf{Progressive Multigrid}\\[1pt]
{\tiny Token $\to$ Chunk $\to$ Section}\\[-1pt]
{\tiny shifted pooling for causality}
\end{tabular}};
\draw[arr] (scanblk.east) -- (mgrid.west);

\node[scan, minimum width=2.0cm] (cross) at (8.2, -1.7) {$\psifield \otimes \psifield$\\{\tiny cross-field}};
\draw[arr] (cross.north) -- ([yshift=-0.15cm]mgrid.south);

\node[mlp, minimum width=2.4cm, minimum height=0.9cm] (rmlp) at (11.4, 0.5) {%
\begin{tabular}{c}
\textbf{Round MLP}\\[1pt]
{\tiny $\mathbf{h} \leftarrow \mathbf{h} + \text{MLP}(\psifield, D, \mathbf{h})$}
\end{tabular}};
\draw[arr] (mgrid.east) -- (rmlp.west);

\node[emb, minimum width=2.0cm] (dec) at (14.8, 0.5) {Decoder\\{\tiny $\to$ logits}};
\draw[arr] (rmlp.east) -- (dec.west);

\draw[arr, blue!50, very thick] (embed.south) -- ++(0,-0.3) -| ([yshift=-0.1cm]rmlp.south);
\node[dimlabel, blue!50] at (7.5, -2.35) {$\mathbf{h}$ residual rail ($D{=}320$)};

\end{tikzpicture}%
}
\caption{\textbf{Causal Poisson language model.} Tokens are embedded and passed through $L{=}6$ non-shared CausalPoissonLayers. Each layer solves the causal Poisson recurrence via $O(N \log N)$ parallel associative scan, applies progressive multigrid (token $\to$ chunk $\to$ section scales with shifted pooling for causal safety), computes cross-field outer products ($\psifield \otimes \psifield$), and integrates all features into the hidden state $\mathbf{h}$ via a round MLP with residual connection.}
\label{fig:lm_arch}
\end{figure}
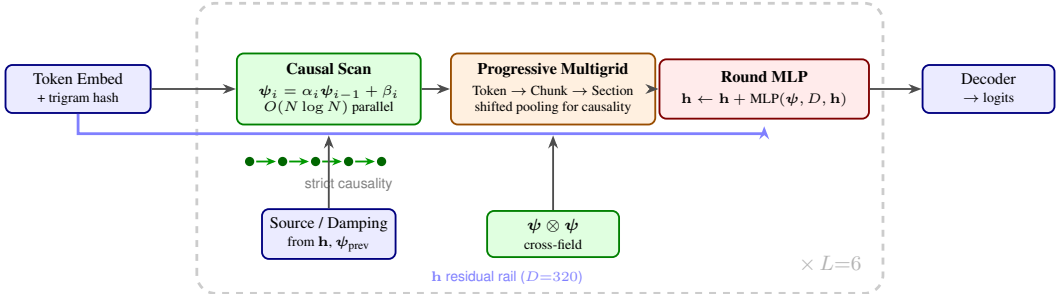

For causal language modeling (Figure~\ref{fig:lm_arch}), the screened Poisson equation is instantiated on a \emph{causal} 1D chain graph: each token position $i$ receives information only from positions $j < i$.
On this causal chain, the Poisson equation reduces to a first-order linear recurrence:
\begin{equation}
    \psifield_i = \alpha_i \cdot \psifield_{i-1} + \beta_i, \quad
    \alpha_i = \frac{w_i}{w_i + \lambda_i}, \quad
    \beta_i = \frac{b_i}{w_i + \lambda_i},
    \label{eq:causal_scan}
\end{equation}
where $w_i$ is a learned edge conductance and $\lambda_i, b_i$ are per-position damping and source.
This affine recurrence is solved exactly in $O(N \log N)$ via the Blelloch parallel associative scan~\citep{blelloch1990prefix}, preserving strict autoregressive causality while providing efficient GPU-parallel execution.

\paragraph{Progressive multigrid.}
A three-level causal hierarchy provides multi-scale context: token-level scans (local syntax), chunk-level scans with shifted pooling (phrase/sentence), and section-level scans (paragraph/topic).
Different layers receive different hierarchy levels---early layers see only local context, later layers access progressively coarser scales.

\paragraph{Cross-field interactions.}
Low-rank outer products $\psifield \otimes \psifield$ and $\psifield \otimes D$ (where $D_i = w_i(\psifield_i - \psifield_{i-1})^2$ is the local dissipation) capture pairwise field correlations, providing the inter-field coupling that attention achieves via multi-head dot products.

\paragraph{Stacked non-shared layers.}
As with CIFAR-100, each of the $L{=}6$ layers has independent weights for source/damping MLPs, edge conductances, and the round MLP that integrates $\psifield$ features into the hidden state $\mathbf{h}$---a design principle shared across all deep Metriplector instantiations.

\section{Experiments}
\label{sec:experiments}

\subsection{CIFAR-100 Image Classification}
\label{sec:cifar100_experiments}

CIFAR-100 provides the strongest test of Metriplector's scalability: 100 fine-grained classes at 32$\times$32 resolution demand both spatial feature extraction and rich cross-class discrimination.
This is the first domain where the full metriplectic structure (Section~\ref{sec:full_metriplectic}) is essential---the Poisson bracket $J$ mediates the cross-field mixing needed to distinguish 100 classes.

\begin{table}[!htb]
\centering
\caption{\textbf{CIFAR-100 results.} All Metriplector models: 12 layers, patch size 2, identical recipe (Adam~\citep{kingma2015adam}, lr=$3{\times}10^{-3}$, cosine, AutoAugment~\citep{cubuk2019autoaugment}, label smoothing 0.1, 200 epochs).}
\label{tab:cifar100}
\small
\begin{tabular}{@{}lccc@{}}
\toprule
\textbf{Model} & \textbf{Params} & \textbf{Acc} & \textbf{Note} \\
\midrule
\multicolumn{4}{@{}l}{\textit{Metriplector (ours, full $M{+}L$)}} \\
\rowcolor{green!10}
\textbf{Metriplectic $K{=}32$ $s{=}8$} & \textbf{2.26M} & \textbf{81.03} & $T^{\mu\nu}$ readout, 8 sub \\
Metriplectic $K{=}32$ $s{=}3$ & 2.26M & 80.6 & 3 substeps \\
Metriplectic $K{=}32$ $s{=}1$ & 2.26M & 80.4 & 1 substep \\
Noether $K{=}32$ & 1.70M & 79.8 & Position readout \\
PhysicsSelects-V2 $K{=}32$ & 1.49M & 78.4 & Curvature readout \\
\midrule
\multicolumn{4}{@{}l}{\textit{Geometric algebra (CliffordNet~\citep{ji2026cliffordnet})}} \\
CliffordNet-64 ($K{=}5$, inner) & 8.6M & 82.46 & Full GA, no FFN \\
CliffordNet-32 ($K{=}3$, full) & 4.8M & 81.42 & Full GA, no FFN \\
CliffordNet-Lite ($K{=}5$, diff) & 2.6M & 79.05 & No FFN \\
CliffordNet-Nano ($K{=}2$, diff) & 1.4M & 77.82 & No FFN \\
\midrule
\multicolumn{4}{@{}l}{\textit{Conventional architectures}} \\
DenseNet-BC~\citep{huang2017densely} & 25.6M & 82.8 & \\
ConvNeXt-T~\citep{liu2022convnet} & 28.6M & 82.0 & \\
WRN-28-10~\citep{zagoruyko2016wide} & 36.5M & 81.1 & \\
ResNet-110~\citep{he2016deep} & 1.7M & 74.1 & \\
\bottomrule
\end{tabular}
\end{table}

\begin{figure}[!htb]
\centering
\includegraphics[width=\textwidth]{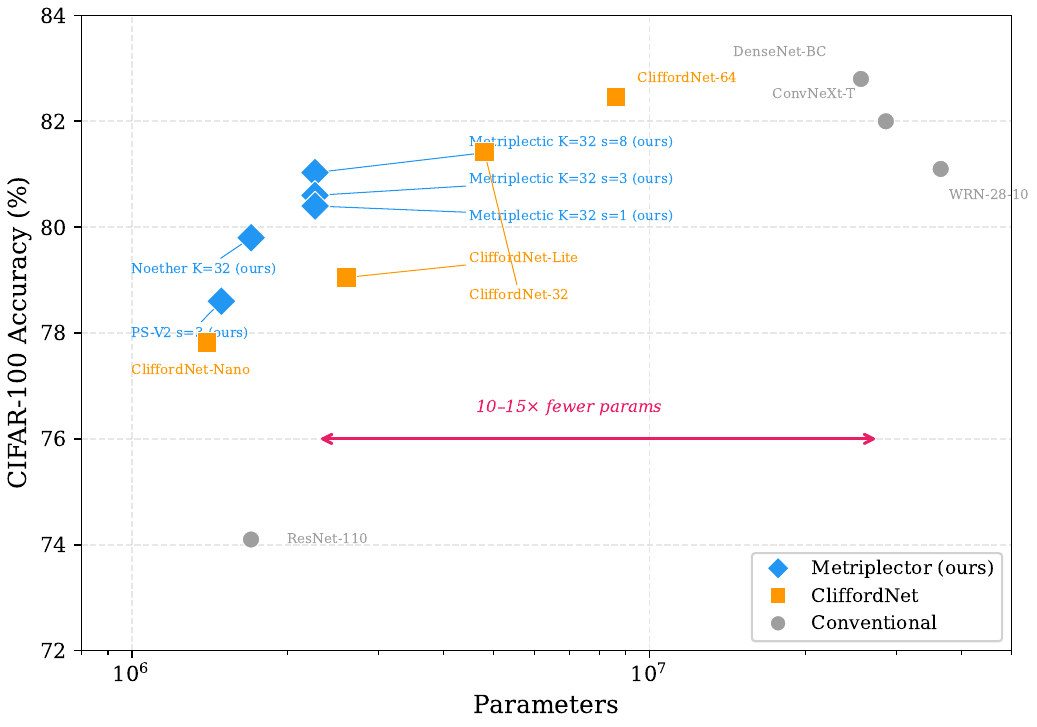}
\caption{\textbf{CIFAR-100: Accuracy vs.\ parameters.} Metriplector variants (blue) achieve 80\%+ accuracy with 2.26M parameters---10--15$\times$ fewer than conventional architectures (gray) at similar accuracy levels.}
\label{fig:params_vs_acc}
\end{figure}

\paragraph{Parameter efficiency.}
Metriplector achieves 81.03\% with only 2.26M parameters---closing to within 1.8 points of DenseNet-BC (82.8\%, 25.6M) while using \textbf{10--15$\times$ fewer parameters}.
Among physics-inspired architectures, the closest comparison is CliffordNet~\citep{ji2026cliffordnet}, which grounds its dynamics in geometric algebra (Clifford product).
At comparable parameter budgets, Metriplector outperforms CliffordNet-Lite (79.05\%, 2.6M) by +2.0 points with fewer parameters, and matches CliffordNet-32 (81.42\%, 4.8M) at less than half the parameters.
On a params-per-accuracy-point basis: WideResNet-28-10 uses 450K params/point, ConvNeXt-Tiny uses 349K, DenseNet-BC uses 309K, while our metriplectic model uses only 28K---an order of magnitude more efficient. This efficiency gap reflects the inductive bias provided by the metriplectic structure. Multiple Euler substeps further improve accuracy (80.4\% $\to$ 80.6\% $\to$ 81.03\% for $s{=}1, 3, 8$), confirming that deeper metriplectic evolution within each layer provides richer spatial information propagation.
This efficiency stems from the $K$-field bottleneck: expensive pairwise interactions occur in a $K{=}32$ dimensional physics space ($K(K{-}1)/2 = 496$ cross-field features) rather than the full $D{=}128$ representation space (Figure~\ref{fig:params_vs_acc}).

\begin{table}[!htb]
\centering
\caption{\textbf{CIFAR-100 ablations.} \textit{Top}: cumulative improvements from V1 (77.6\%); each row adds one component on top of all previous rows; $\Delta$ is relative to V1. \textit{Bottom}: component removal from V2 baseline (78.4\%); $\Delta$ is the accuracy drop when that component is removed.}
\label{tab:cifar100_ablations}
\begin{tabular}{lcc}
\toprule
\textbf{Ablation} & \textbf{Acc (\%)} & \textbf{$\Delta$ from base} \\
\midrule
\multicolumn{3}{@{}l}{\textit{Cumulative improvements (from V1 = 77.6\%)}} \\
\quad + SiLU + BN residual engineering (V1 $\to$ V2) & 78.4 & +0.8 \\
\quad + 3 Euler substeps (more propagation per layer) & 78.6 & +1.0 \\
\quad + Noether currents ($p_x, p_y, L, D$, position-aware) & 79.8 & +2.2 \\
\quad + Stress-energy + 8 substeps ($\nabla\psifield{\times}\nabla\psifield$, $K^2$ features) & 81.03 & +3.4 \\
\midrule
\multicolumn{3}{@{}l}{\textit{Component removal (from V2 baseline = 78.4\%)}} \\
\quad $-$ $K{=}32$ fields ($\to$ $K{=}16$) & 74.5 & $-$3.9 \\
\quad $-$ Fresh $\psifield$ per layer ($\to$ recurrent $\psifield$) & 75.0 & $-$3.4 \\
\quad $-$ Operator from $\mathbf{h}$ ($\to$ operator from $\psifield$) & 64.1 & $-$14.3 \\
\quad $-$ Poisson bracket $J$ ($\to$ diffusion only) & 65.0 & $-$13.4 \\
\bottomrule
\end{tabular}
\end{table}

\begin{figure}[!htb]
\centering
\includegraphics[width=\textwidth]{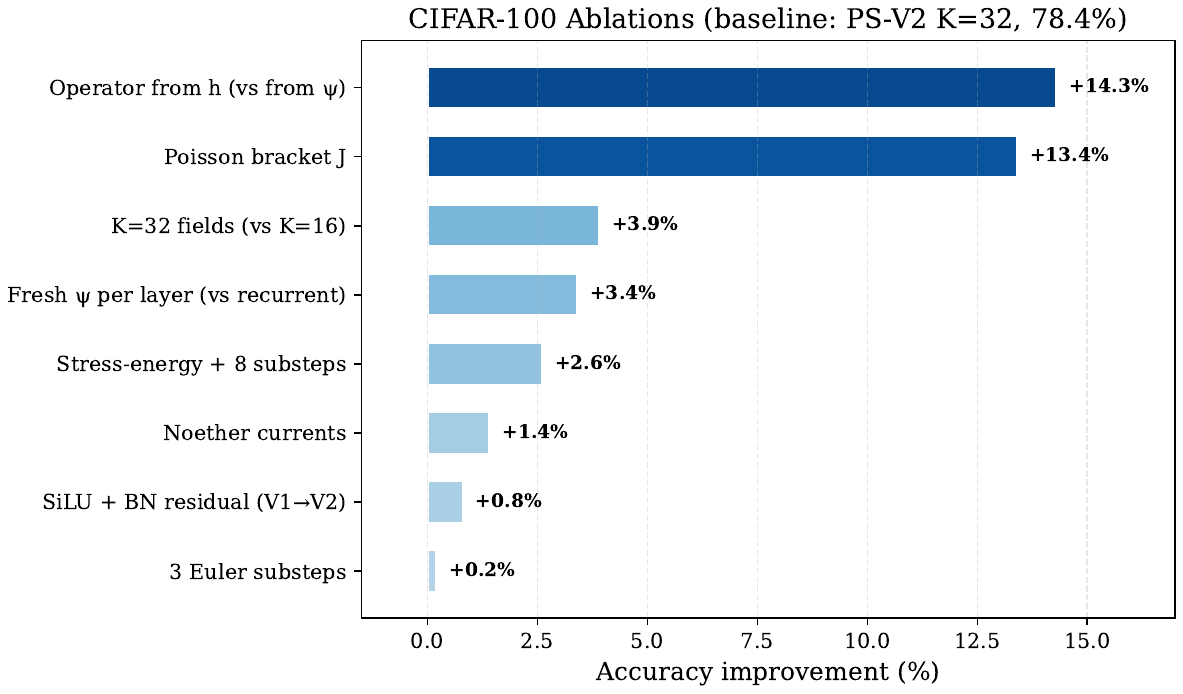}
\caption{\textbf{CIFAR-100 ablation impacts.} Accuracy improvement from each architectural component relative to the PhysicsSelects-V2 $K{=}32$ baseline (78.4\%). The operator-from-$\mathbf{h}$ principle (+14.3\%) and the Poisson bracket $J$ (+13.4\%) are the most important factors. Fresh $\psifield$ per layer (+3.4\%) validates the non-recurrent physics design.}
\label{fig:ablations}
\end{figure}

The ablations confirm that both branches of the metriplectic formalism are essential: removing the Poisson bracket $J$ (the antisymmetric cross-field coupling that mediates advection) costs \textbf{13.4 points}.
The stress-energy tensor readout---derived from Noether's theorem connecting spatial symmetries to conserved quantities---provides the most effective feature extraction (Figure~\ref{fig:ablations}).

\begin{figure}[!htb]
\centering
\includegraphics[width=\textwidth]{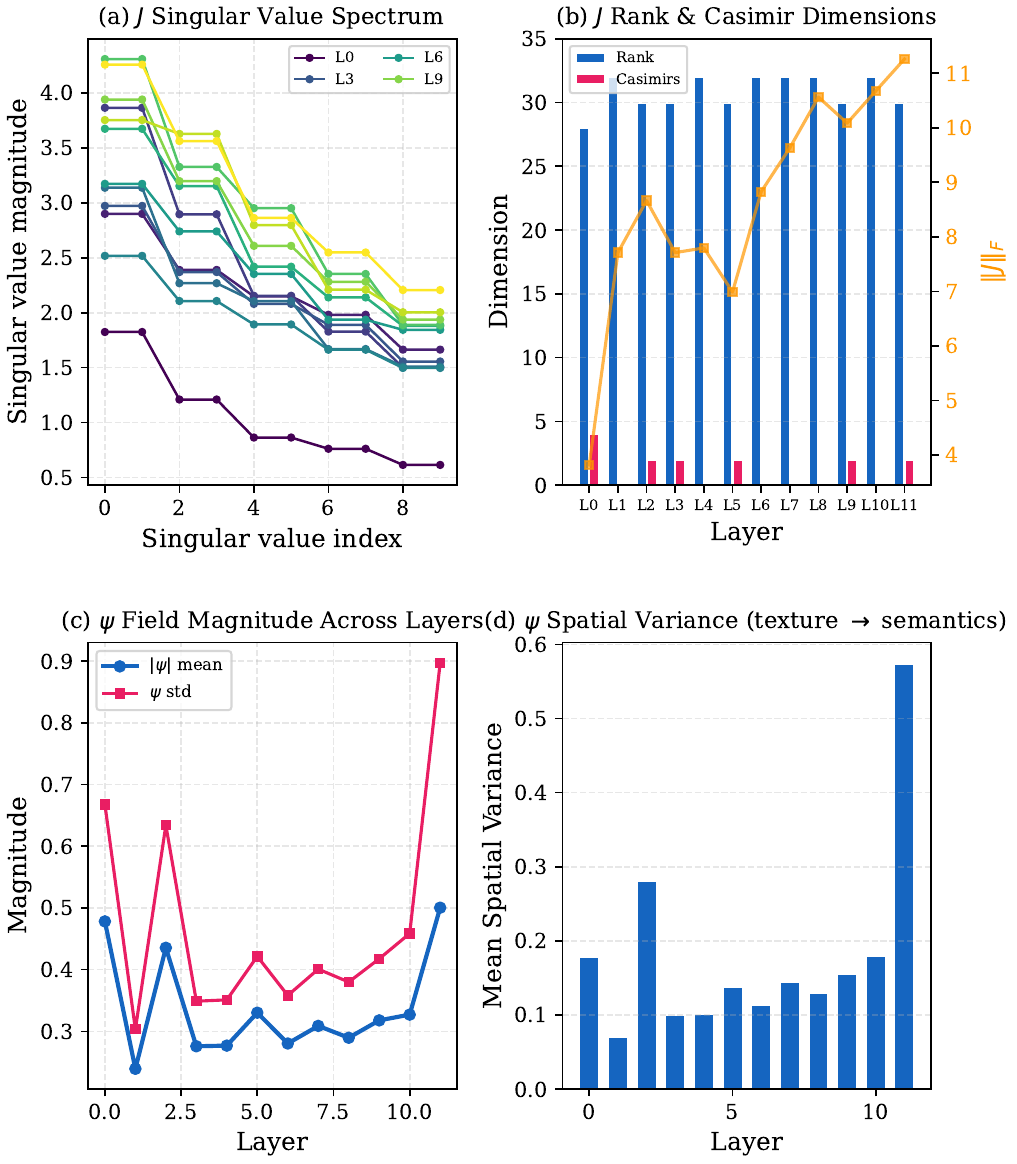}
\caption{\textbf{CIFAR-100 physics diagnostics across 12 layers.} \emph{(a)} Singular value spectrum of the learned Poisson tensor $J$---values appear in degenerate pairs, the hallmark of skew-symmetric structure. \emph{(b)} $J$ effective rank and null-space dimensions; $\|J\|_F$ grows monotonically with depth, indicating stronger cross-field coupling in later layers. \emph{(c)} $\psifield$ magnitude and standard deviation remain stable (0.25--0.5) across L0--L10, with a sharp spike at L11 before the stress-energy readout. \emph{(d)} Spatial variance increases at the final layer, reflecting transition from distributed texture to localized semantic features.}
\label{fig:cifar_diagnostics}
\end{figure}

\paragraph{Physics diagnostics.}
Figure~\ref{fig:cifar_diagnostics} reveals the learned Poisson tensor $J$ across layers: singular values appear in degenerate pairs (a structural signature of skew-symmetry), and the effective rank grows with depth while the null-space dimensions reflect the intrinsic rank deficiency of skew-symmetric matrices.
The $J$ Frobenius norm grows from 3.8 (L0) to 10.9 (L11), suggesting deeper layers develop stronger cross-field coupling.
The $\psifield$ field magnitude and spatial variance across layers show a sharp increase at L11, reflecting the final layer concentrating discriminative features before the stress-energy readout.

\begin{figure}[!htb]
\centering
\includegraphics[width=\textwidth]{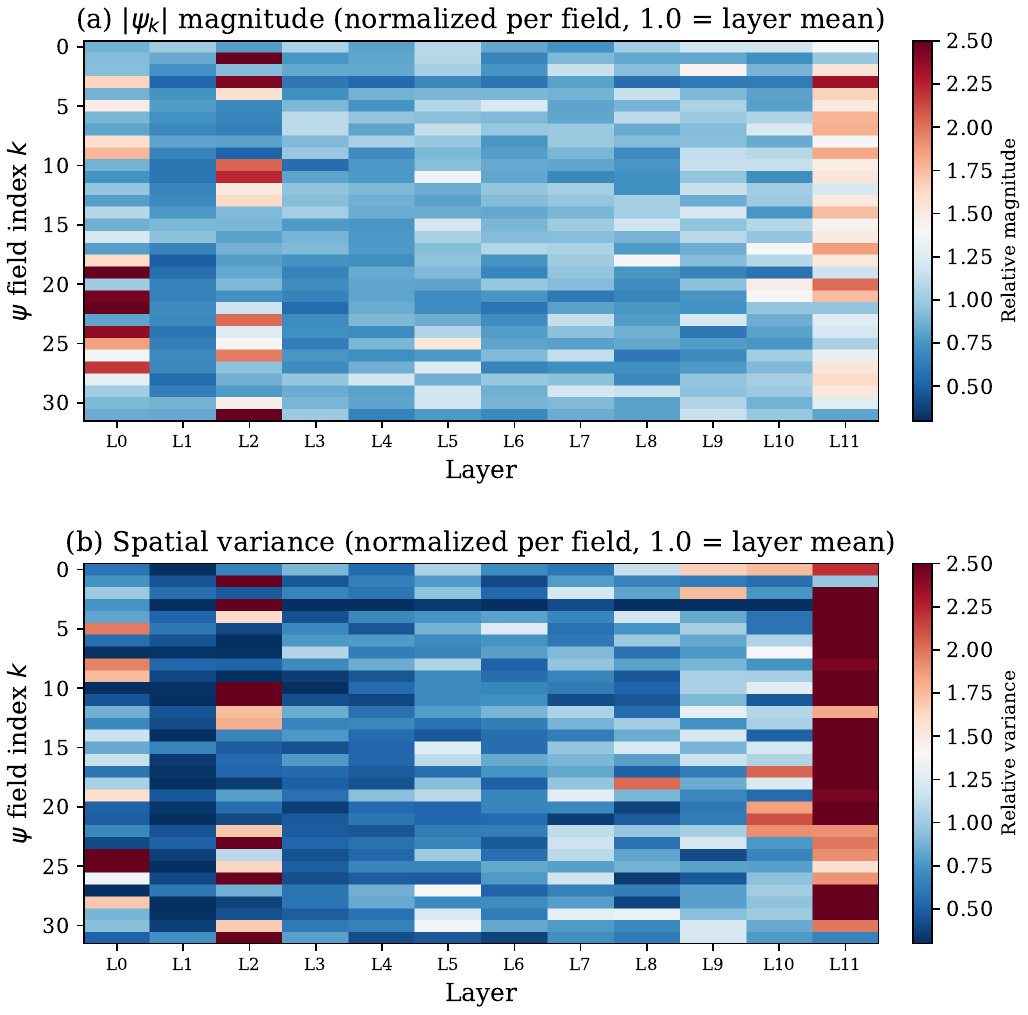}
\caption{\textbf{$\psifield$ field specialization across 12 layers} (32 fields $\times$ 12 layers). \emph{(a)} Field magnitude: most fields remain moderate through L0--L10, with a sharp spike at L11 where specific fields (e.g., $k{=}3, 20, 21$) concentrate discriminative features. \emph{(b)} Spatial variance: early layers develop spatially uniform fields (low variance); L11 shows dramatic spatial structure as the model localizes class-specific features before readout.}
\label{fig:cifar_field_heatmap}
\end{figure}

\paragraph{Field specialization and dynamics budget.}
Figure~\ref{fig:cifar_field_heatmap} reveals how the 32 $\psifield$ fields specialize across layers.
Field magnitudes remain moderate (0.2--0.5) through L0--L10, with a dramatic spike at L11 where several fields exceed magnitude 1.0---the model concentrates discriminative features into specific field channels before the stress-energy readout.
Spatial variance follows the same pattern: most fields maintain low variance (spatially uniform activations) in early layers, developing strong spatial structure only at L11 where the network must localize class-specific features.
This pattern---stable evolution followed by a concentrated burst---is consistent with the metriplectic dynamics operating as a controlled energy cascade.

\begin{figure}[!htb]
\centering
\includegraphics[width=\textwidth]{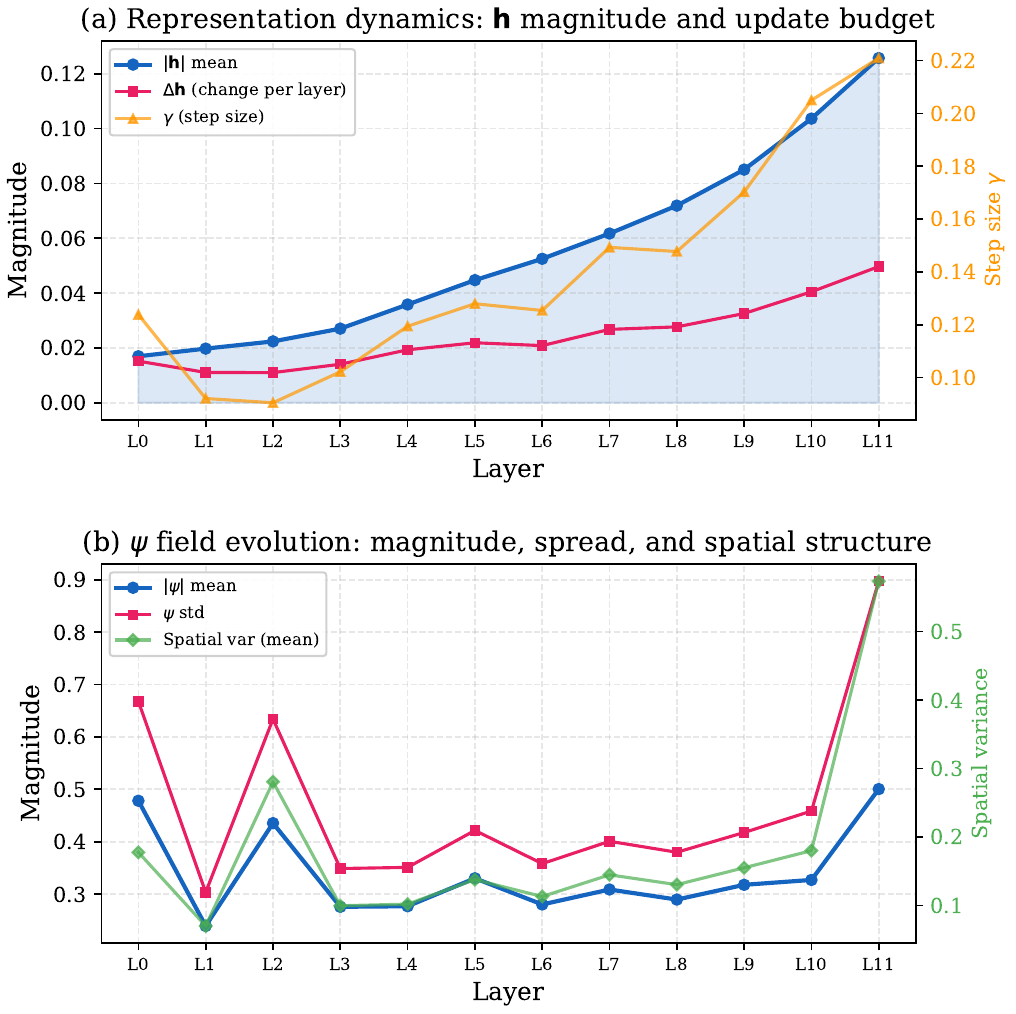}
\caption{\textbf{CIFAR-100 forward dynamics budget.} \emph{(a)} Hidden state $\mathbf{h}$: magnitude and per-layer update $\Delta\mathbf{h}$ both grow monotonically; the step size $\gamma$ increases from 0.12 to 0.22, self-organizing an expanding dynamics budget. \emph{(b)} $\psifield$ fields: magnitude and standard deviation are stable (0.25--0.5) across L0--L10, with spatial variance spiking at L11.}
\label{fig:cifar_dynamics}
\end{figure}

Figure~\ref{fig:cifar_dynamics} shows the representation dynamics budget across layers: the hidden state $\mathbf{h}$ grows monotonically from $|\mathbf{h}| = 0.017$ (L0) to $0.126$ (L11), with each layer contributing a controlled update $\Delta\mathbf{h}$ that grows proportionally.
The learned step size $\gamma$ increases with depth (0.12 $\to$ 0.22), allowing later layers to make larger representational changes---the architecture self-organizes an expanding dynamics budget.

\FloatBarrier
\subsection{Reacher Robotic Control}
\label{sec:reacher_experiments}

The Reacher domain tests whether metriplectic dynamics can serve as a latent-space world model for robotic control, competing with transformer-based architectures that are an order of magnitude larger.
A 2-joint, 6-DOF robotic arm must reach a target joint configuration; the world model is trained on 10K episodes of random-policy trajectories (64$\times$64 pixel observations), and evaluated via CEM planning with the \texttt{qpos\_match} criterion (all joint angles within 0.05 radians of target)---identical to the LeWM~\citep{maes2024lewm} evaluation protocol.

\begin{table}[!htb]
\centering
\caption{\textbf{Reacher: comparison with baselines.} All methods use the \texttt{qpos\_match} criterion from~\citet{maes2024lewm}. CEM: 50 episodes, 300 candidates, 30 elite, 30 iterations, horizon~5.}
\label{tab:reacher_baselines}
\small
\begin{tabular}{@{}lccp{4.2cm}@{}}
\toprule
\textbf{Model} & \textbf{Params} & \textbf{CEM} & \textbf{Architecture} \\
\midrule
LeWM & $\sim$15M & 86\% & ViT-Tiny + 6-layer transformer \\
DINO-WM & $>$21M & 79\% & Frozen DINOv2 ViT-S \\
PLDM & $\sim$15M & 78\% & ViT + GRU predictor \\
Random & --- & 10\% & --- \\
\midrule
\rowcolor{green!10}
\textbf{Ours} & \textbf{913K} & \textbf{88\%} & 4$\times$ MetriplectorLayer \\
\bottomrule
\end{tabular}
\end{table}

\paragraph{Parameter efficiency.}
Our metriplectic world model uses 913K parameters and reaches 88\% CEM success---comparable to the strongest baseline (LeWM, 86\%) while using 16$\times$ fewer parameters (Table~\ref{tab:reacher_baselines}).
The convolutional encoder ($\sim$560K params) and 4 MetriplectorLayers ($\sim$353K params) achieve this with an order of magnitude fewer parameters than transformer-based world models that rely on pre-trained vision encoders or large causal transformers.
With 50 evaluation episodes, the 2-point margin is within statistical noise; the significant finding is that metriplectic dynamics matches transformer-scale performance at a fraction of the capacity.

\paragraph{Symplectic structure and autoregressive training.}
Two innovations drive the improvement from 82\% (v4.7, teacher-forcing only) to 88\% (v5.1):
(1)~the canonical symplectic $J_\Omega$ organizes the $K{=}16$ fields into 8~conjugate $(q,p)$ pairs, providing built-in Hamiltonian coupling that stabilizes field dynamics;
(2)~multi-step AR training (activated at epoch~10 after teacher-forcing warmup) reduces step-5 MSE by 4$\times$, directly improving multi-step CEM rollout quality.
The model demonstrates strong controllable dynamics: \texttt{act\_ratio} = 5.3$\times$ (the model strongly distinguishes different action sequences), prediction loss decreases monotonically (0.087$\to$0.032 step-1 MSE), latent norms remain perfectly stable across 5-step autoregressive rollouts (80$\to$78$\to$77$\to$76$\to$75, only 6\% decay), and 95\% of latent dimensions remain active with zero dead channels.

\begin{figure}[!htb]
\centering
\includegraphics[width=\linewidth]{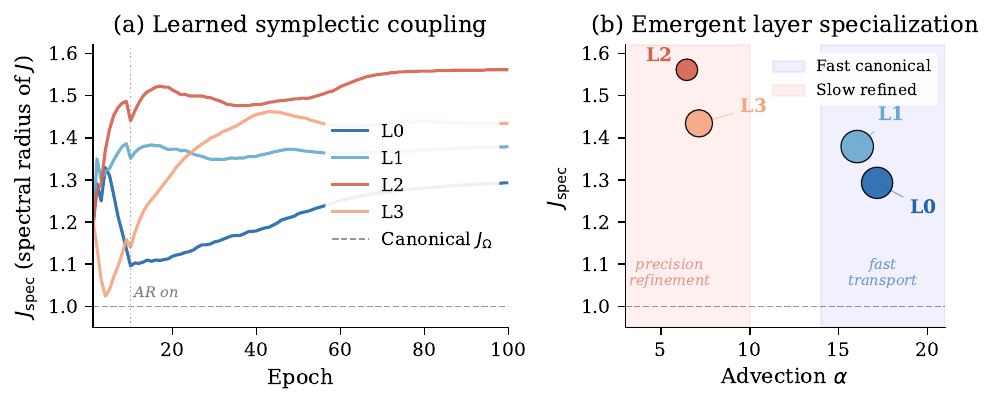}
\caption{\textbf{Emergent symplectic layer specialization.} (a)~Evolution of $J_\mathrm{spec}$ (spectral radius of the full Poisson bracket $J = J_\Omega + J_\mathrm{cross}$) during training. All layers begin near canonical ($J_\mathrm{spec} = 1.0$) and develop a depth-ordered hierarchy: L2 acquires the richest inter-pair coupling ($+$56\% beyond canonical). The ``AR on'' marker at epoch~10 shows the onset of multi-step autoregressive training. (b)~Final layer configuration: advection strength $\alpha$ vs.\ $J_\mathrm{spec}$ reveals two emergent physical regimes---``fast canonical transport'' (L0--L1, high $\alpha$) and ``slow precision refinement'' (L2--L3, low $\alpha$, high coupling)---resembling symplectic splitting integrators~\citep{yoshida1990construction}. Bubble size $\propto$ energy flux $|dH|$.}
\label{fig:reacher_jspec}
\end{figure}

\paragraph{Emergent physics from learned dynamics.}
The per-layer diagnostics reveal physically meaningful structure that emerges from training without architectural enforcement.
The spectral radius of the full Poisson bracket $J = J_\Omega + J_\text{cross}$---which equals 1.0 for canonical structure alone---develops a clear depth hierarchy: layers L0--L1 remain near canonical ($J_\text{spec} \approx 1.3$, $+$30\% learned coupling), while layer L2 develops the richest inter-pair interactions ($J_\text{spec} = 1.56$, $+$56\% beyond canonical; Figure~\ref{fig:reacher_jspec}a).
This correlates with a functional specialization: L0--L1 operate as ``fast canonical transport'' layers (advection $\alpha \approx 17$, large updates), while L2 acts as a ``precision refinement'' layer (advection $\alpha \approx 6.5$, smallest gate and smallest update magnitude; Figure~\ref{fig:reacher_jspec}b).
The network independently discovers a strategy resembling symplectic splitting methods~\citep{yoshida1990construction}: fast evolution under the canonical Hamiltonian first, followed by slow, precise corrections from learned inter-pair couplings.
Because both $J_\Omega$ (fixed) and $J_\text{cross} = W - W^\top$ (learned) are antisymmetric by construction, the advection term $\alpha J \psifield$ generates volume-preserving flow at every layer---a discrete instantiation of Liouville's theorem.
The 6\% norm decay over 5-step rollouts comes entirely from the dissipative GENERIC branch (diffusion and damping), confirming the intended separation of reversible and irreversible dynamics.

\FloatBarrier
\subsection{Sudoku Constraint Satisfaction}

Metriplector achieves \textbf{99.4\%} cell accuracy and \textbf{97.2\%} exact solve rate on 9$\times$9 Sudoku with \emph{zero structural injection}: the graph has only 8-connected spatial edges (544 total), and no row, column, or box structure is provided (120K parameters, 10K training puzzles, $R{=}32$ rounds).
Per difficulty: easy 99.8\%, medium 99.6\%, hard 99.0\%.
The multigrid object layer is the largest single contributor (+10\% cell accuracy); increasing rounds from 16 to 32 adds +4.8\%; scaling from 500 to 10K training puzzles adds +6.4\%.
Data augmentation (digit permutation, spatial rotations) consistently hurts.

\paragraph{Emergent box discovery.}

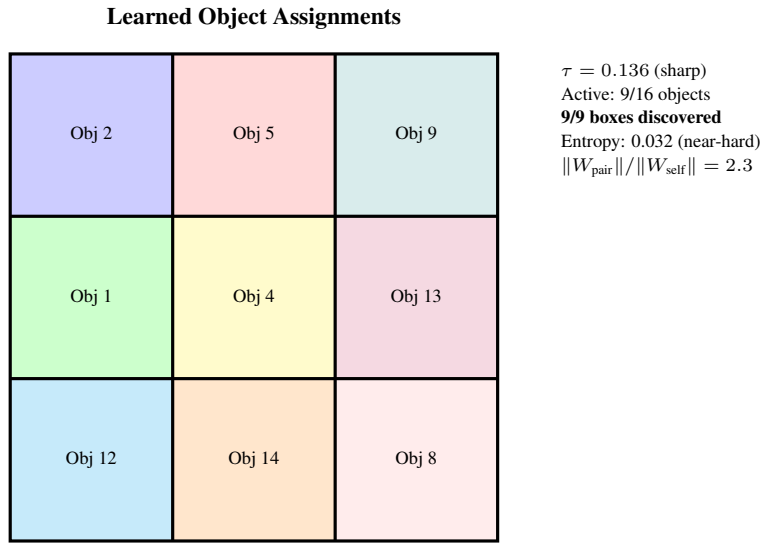
\begin{figure}[!htb]
\centering
\begin{tikzpicture}[scale=0.72]
\draw[gray!30] (0,0) grid (9,9);
\fill[blue!20] (0,6) rectangle (3,9);
\node[font=\scriptsize] at (1.5,7.5) {Obj 2};
\fill[red!15] (3,6) rectangle (6,9);
\node[font=\scriptsize] at (4.5,7.5) {Obj 5};
\fill[teal!15] (6,6) rectangle (9,9);
\node[font=\scriptsize] at (7.5,7.5) {Obj 9};
\fill[green!20] (0,3) rectangle (3,6);
\node[font=\scriptsize] at (1.5,4.5) {Obj 1};
\fill[yellow!25] (3,3) rectangle (6,6);
\node[font=\scriptsize] at (4.5,4.5) {Obj 4};
\fill[purple!15] (6,3) rectangle (9,6);
\node[font=\scriptsize] at (7.5,4.5) {Obj 13};
\fill[cyan!20] (0,0) rectangle (3,3);
\node[font=\scriptsize] at (1.5,1.5) {Obj 12};
\fill[orange!20] (3,0) rectangle (6,3);
\node[font=\scriptsize] at (4.5,1.5) {Obj 14};
\fill[pink!30] (6,0) rectangle (9,3);
\node[font=\scriptsize] at (7.5,1.5) {Obj 8};
\draw[very thick] (0,0) rectangle (9,9);
\draw[very thick] (3,0) -- (3,9);
\draw[very thick] (6,0) -- (6,9);
\draw[very thick] (0,3) -- (9,3);
\draw[very thick] (0,6) -- (9,6);
\node[font=\small\bfseries, anchor=south] at (4.5,9.3) {Learned Object Assignments};
\node[font=\scriptsize, align=left, anchor=north west] at (10,9) {
    $\tau = 0.136$ (sharp)\\[1pt]
    Active: 9/16 objects\\[1pt]
    \textbf{9/9 boxes discovered}\\[1pt]
    Entropy: 0.032 (near-hard)\\[1pt]
    $\|W_\text{pair}\|/\|W_\text{self}\| = 2.3$
};
\end{tikzpicture}
\caption{\textbf{Emergent box discovery in Sudoku.} The object layer discovers all 9 Sudoku 3$\times$3 boxes from an 8-connected spatial lattice with no box-level supervision. Assignments are sharp ($\tau{=}0.136$, entropy ${\approx}0.032$) and static across all puzzles.}
\label{fig:boxes}
\end{figure}

Post-hoc analysis reveals that the object layer discovers \textbf{all 9} Sudoku 3$\times$3 boxes from the 8-connected spatial lattice (Figure~\ref{fig:boxes}).
Assignment temperature $\tau{=}0.136$ produces near-deterministic groupings (mean entropy 0.032), with exactly 9 of 16 available objects active---each containing exactly 9 cells aligned to box boundaries.
The cross-object interaction ratio $\|W_\text{pair}\|/\|W_\text{self}\| = 2.3$ indicates the model prioritizes inter-box communication.
Directional scans handle row/column constraints while objects handle box constraints---this division of labor emerges from training without any structural hints.

\begin{figure}[!htb]
\centering
\includegraphics[width=0.5\textwidth]{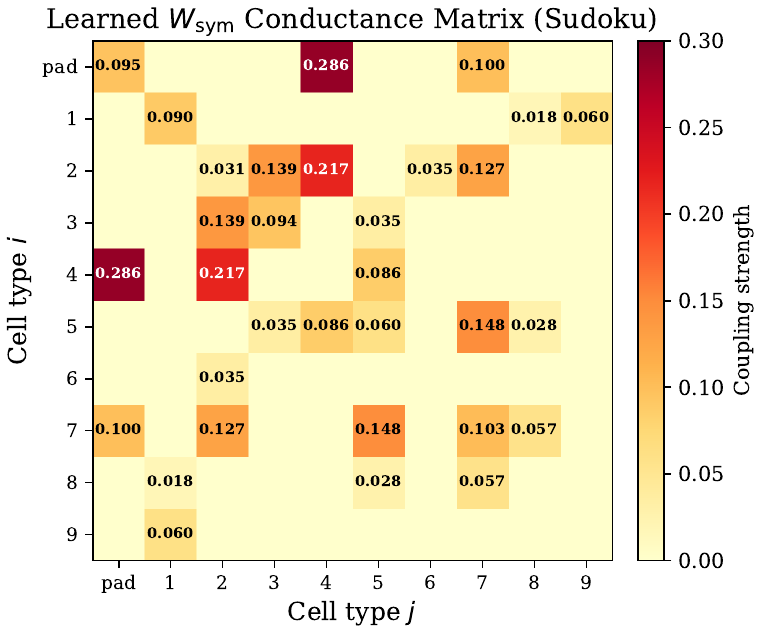}
\caption{\textbf{Learned conductance matrix} $W_\text{sym}$ (Sudoku). Entries show coupling strength between cell types (digits 1--9 and padding). The model learns sparse, structured couplings: strong links (4$\leftrightarrow$pad, 2$\leftrightarrow$4, 5$\leftrightarrow$7) emerge without supervision.}
\label{fig:sudoku_wsym}
\end{figure}

\FloatBarrier
\subsection{Language Modeling}
\label{sec:lm_experiments}

Language modeling tests whether the metriplectic formalism can compete with transformers on their home turf: autoregressive next-token prediction on the FineWeb dataset~\citep{penedo2024fineweb} with a 1024-token SentencePiece vocabulary, measuring bits per byte (BPB, lower is better).
All models respect the $\leq$16\,MB artifact size constraint of the Parameter Golf challenge.

\begin{table}[!htb]
\centering
\caption{\textbf{Language modeling results.} BPB on FineWeb validation. All models $\leq$16\,MB.}
\label{tab:lm}
\small
\begin{tabular}{lcccl}
\toprule
\textbf{Model} & \textbf{Params} & \textbf{BPB $\downarrow$} & \textbf{Tokens} & \textbf{Note} \\
\midrule
\rowcolor{green!10}
\textbf{Metriplector (Stacked v6)} & \textbf{18.5M} & \textbf{1.182} & 2.0B & 1$\times$GH200, 2.6\,hr \\
GPT baseline & 9.8M & 1.224 & 7.2B & 8$\times$H100, 10\,min \\
PG leaderboard SOTA & --- & 1.119 & --- & +TTT, EMA, etc. \\
\bottomrule
\end{tabular}
\end{table}

The Stacked v6 model uses 6 non-shared CausalPoissonLayers with progressive multigrid and cross-field outer products ($\psifield \otimes \psifield$), trained for 30K steps on 512-token sequences.
It achieves \textbf{1.182 BPB}, compared to the GPT baseline (1.224 BPB, post-quantization)---a difference of 0.042 BPB while seeing \textbf{3.6$\times$ fewer training tokens} (2.0B vs.\ 7.2B).
The comparison is not compute-matched: GPT trains on 8 GPUs in 10 minutes (80 GPU-minutes) while our model uses a single GPU for 2.6 hours (156 GPU-minutes, $\sim$2$\times$ more).
Our model uses 512-token sequences vs.\ GPT's 1024; BPB is normalized per byte, but longer-context modeling is inherently harder.
The PG leaderboard SOTA (1.119 BPB) exceeds both models using techniques orthogonal to architecture.

\begin{figure}[!htb]
\centering
\includegraphics[width=\textwidth]{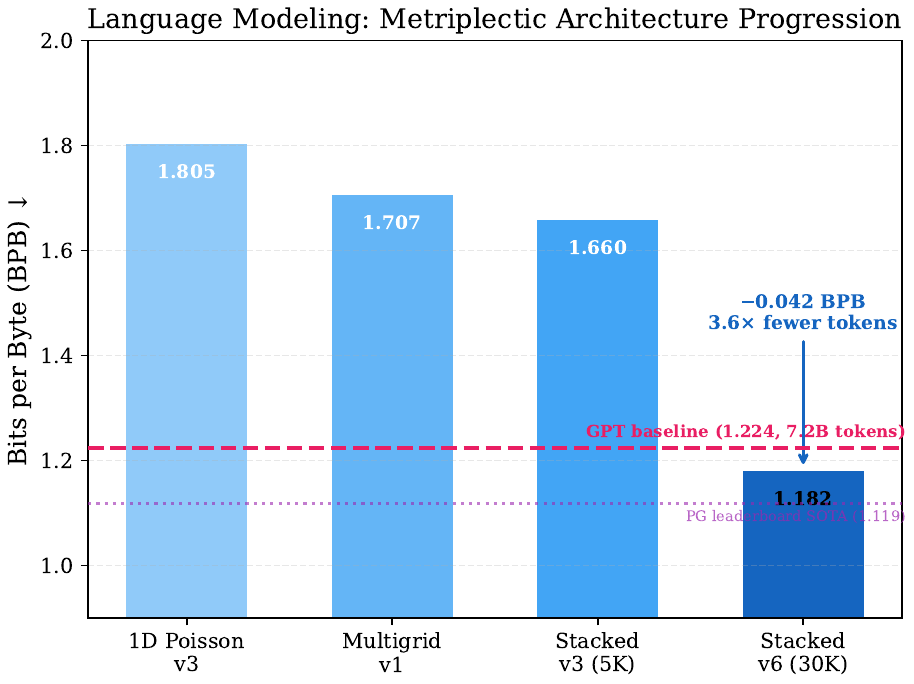}
\caption{\textbf{Language modeling BPB progression.} GPT baseline (1.224 BPB, dashed) shown for reference; Metriplector reaches 1.182 BPB with 3.6$\times$ fewer training tokens.}
\label{fig:lm_bpb}
\end{figure}

\FloatBarrier
\subsection{Maze Pathfinding}

\begin{table}[!htb]
\centering
\caption{\textbf{Maze results.} F1 score on tree-maze path prediction. Metriplector achieves F1\,=\,1.0 on 15$\times$15 and F1\,=\,0.95--0.99 on 39$\times$39 across seeds.}
\label{tab:maze}
\begin{tabular}{lcccc}
\toprule
\textbf{Model} & \textbf{Params} & \textbf{F1 (15$\times$15)} & \textbf{F1 (39$\times$39)} & \textbf{Steps} \\
\midrule
Harmonic baseline & 0 & 0.47 & --- & --- \\
MetriNet Phase 7.3B & 9,100 & 0.59 & --- & 4,000 \\
Metriplector 7.53K ($K{=}2$) & 43.8K & 1.00 & 0.43 & 10,000 \\
\rowcolor{green!10}
\textbf{Metriplector 7.53O ($\lambda/N$)} & \textbf{43.8K} & \textbf{1.00} & \textbf{0.95--0.99} & \textbf{10,000} \\
\bottomrule
\end{tabular}
\end{table}

The maze domain demonstrates Metriplector's most striking result: \textbf{robust size generalization}.
Training on 250 mazes of size 15$\times$15 and evaluating on 200 mazes of size 39$\times$39 (6.8$\times$ more cells, 17$\times$ longer paths), the model achieves F1\,=\,1.0 on 15$\times$15 and F1\,=\,0.95--0.99 on 39$\times$39 across seeds (43.8K parameters).
Our $\lambda/N$ scaling preserves the spectral balance of the graph Laplacian across grid sizes.

\paragraph{Emergent type discovery.}
The learned conductance matrix $W \in \R^{4 \times 4}$ discovers that one cell type (walls) should have near-zero conductance while another (corridors) should conduct freely.
The source encoder learns to inject positive charge at the source and negative at the goal.
All of this emerges from the anonymous type principle---the model receives only integer indices, never semantic labels.

\FloatBarrier
\subsection{Cross-Domain Summary}

\begin{table}[!htb]
\centering
\caption{\textbf{Cross-domain results.} One metriplectic formalism, five instantiations.}
\label{tab:summary}
\small
\begin{tabular}{lccl}
\toprule
\textbf{Domain} & \textbf{Params} & \textbf{Result} & \textbf{Key Finding} \\
\midrule
CIFAR-100 & 2.26M & 81.03\% acc & 10--15$\times$ fewer params vs.\ DenseNet \\
Reacher & 913K & 88\% CEM & 16$\times$ fewer params than baselines \\
Sudoku 9$\times$9 & 120K & 97.2\% exact & 9/9 box discovery, zero injection \\
Language & 18.5M & 1.182 BPB & 3.6$\times$ fewer tokens vs.\ GPT \\
Maze 15$\to$39 & 43.8K & F1\,=\,0.95--0.99 & Robust size transfer \\
\bottomrule
\end{tabular}
\end{table}

\noindent A single metriplectic formalism, instantiated differently per domain, achieves competitive results across all five tasks with consistently fewer parameters and training samples than conventional approaches (Table~\ref{tab:summary}).
The Reacher result is particularly significant: it demonstrates that the same $T^{\mu\nu}$ readout and Euler-based metriplectic dynamics used for image classification can serve as a latent-space world model for robotic control, reaching 88\% CEM success with 16$\times$ fewer parameters than the strongest baselines (913K vs.\ $\sim$15M).

\section{Discussion}
\label{sec:discussion}

\paragraph{Physics and learning are complementary.}
Metriplector is not purely physics-based.
The Poisson fields provide \emph{spatial propagation}; the MLPs (cell encoder, damping, source, decoder) perform \emph{learned reasoning}.
Neither alone suffices: pure Poisson reaches $\sim$68\% on Sudoku; adding learned components reaches 99.4\% cell accuracy.
Across all five domains, this combination yields parameter and sample efficiency: 10--15$\times$ fewer params/accuracy-point on CIFAR-100, robust size generalization on maze (F1\,$>$\,0.95 at 6.8$\times$ grid scale), 3.6$\times$ fewer training tokens for language modeling, and over an order of magnitude parameter reduction on Reacher world modeling.

\paragraph{The operator-from-input principle.}
The most consequential design decision, inherited from the maze solver, is that $\mathbf{h}$ defines the operator and $\psifield$ is the solution.
In early experiments, architectures where $\psifield$ built its own operator suffered from field collapse ($\psifield$ magnitude $<$1\% of $\mathbf{h}$ by layer 12).
Ablations confirm the importance: violating this principle costs 14.3 accuracy points on CIFAR-100 (Table~\ref{tab:cifar100_ablations}).

\paragraph{The Poisson bracket is essential.}
Removing the antisymmetric Poisson bracket $J$ costs 13.4 accuracy points (78.4\% $\to$ 65.0\%).
Diffusion alone ($M$ only) builds spatial coherence but not the cross-field interactions needed to distinguish 100 fine-grained categories.
The conservative advection channel provides the missing feature mixing at $O(K^2)$ cost.

\paragraph{Symplectic structure and Liouville stability in world modeling.}
The Reacher domain reveals how physical structure---not just architectural engineering---enables stable long-horizon prediction with far fewer parameters.
The central mechanism is the canonical symplectic Poisson bracket $J_\Omega$, which organizes the $K$ PDE fields as conjugate $(q_i, p_i)$ pairs.
Because $J_\Omega$ is antisymmetric by construction, the advection term $\alpha J_\Omega \psifield$ generates \emph{volume-preserving} flow on the field manifold---a direct instantiation of Liouville's theorem.
This provides a structural guarantee that the reversible dynamics cannot expand or contract the latent representation, regardless of what parameters the network learns.
The learned cross-pair coupling $J_\text{cross} = W - W^\top$, also antisymmetric by parametrization, extends the Hamiltonian structure to inter-pair interactions while preserving the same volume invariant.

The diagnostics confirm that this physical structure produces measurable effects: latent norms decay only 6\% over 5-step autoregressive rollouts (80$\to$75), with the decay attributable entirely to the dissipative GENERIC branch (diffusion and damping)---exactly the intended separation of reversible and irreversible dynamics.
The four layers self-organize into two distinct physical regimes (Figure~\ref{fig:reacher_jspec}): ``fast canonical transport'' (L0--L1, advection $\alpha \approx 17$, near-canonical $J_\text{spec} \approx 1.3$) and ``slow precision refinement'' (L2, $\alpha \approx 6.5$, richest learned coupling $J_\text{spec} = 1.56$), resembling the splitting strategies used in symplectic integrators~\citep{yoshida1990construction}.
None of this specialization is architecturally enforced---it emerges from the combination of canonical symplectic structure and gradient-based training.

The port-Hamiltonian action conditioning completes the physics: the action modulates the PDE operators ($\sigma, \alpha, \gamma, \mathbf{s}$) via zero-initialized projections, instantiating the $G \cdot \mathbf{u}$ port term from port-Hamiltonian control theory.
Because the action configures the \emph{operators} rather than modifying the \emph{state}, the latent representation passes through unchanged, and the volume-preservation guarantee of $J$ extends seamlessly to action-conditioned multi-step rollouts.

\paragraph{Architectural unification: dissipation is stress-energy, assignment is renormalization.}
We now formalize the claim made in the introduction: the Sudoku and CIFAR-100 instantiations implement the \emph{same physics} under different names.

In the Sudoku architecture, the ``dissipation'' feature readout computes
\begin{equation}
    D_k(i) = \sum_j w_{ij}\bigl(\psifield_k(i) - \psifield_k(j)\bigr)^2,
    \label{eq:dissipation_is_Tmunu}
\end{equation}
which is the discrete approximation to $|\nabla \psifield_k|^2$---\emph{exactly} the diagonal of the stress-energy tensor $T_{kk}^{xx} + T_{kk}^{yy} = (\partial_x\psifield_k)^2 + (\partial_y\psifield_k)^2$ that the CIFAR-100 architecture computes via learned gradient convolutions.
The gradient correlation $\nabla\psifield_a \cdot \nabla\psifield_b$ is the same computation as the structure tensor, which has been effective for edge and corner detection in classical computer vision since the 1980s~\citep{harris1988combined,forstner1986feature}. Applied to learned fields evolved via metriplectic dynamics rather than raw image gradients, it arises from Noether's theorem as the conserved quantity of spatial translation symmetry---and generalizes far beyond vision: the same readout principle provides effective feature extraction for image classification, constraint satisfaction, language modeling, and robotic control across all domains tested.
The CIFAR variant additionally extracts the full off-diagonal terms $E_{ab} = \nabla\psifield_a \cdot \nabla\psifield_b$ and vorticity $V_{ab} = \nabla\psifield_a \times \nabla\psifield_b$; extending the Sudoku readout to the full $T^{\mu\nu}$ is a natural next step.

The multigrid object layer reveals a structural analogy with block-spin coarse-graining~\citep{kadanoff1966scaling}.
The soft assignment $\rho_k(i) = \softmax\bigl(\text{MLP}(\psifield, \mathbf{p})/\tau\bigr)$ with coarse-graining $\psifield^{(1)} = \rho^\top \psifield^{(0)}$ averages fine-grained fields into learned groups, an effective dynamics operates at the coarse scale, and the result is prolongated back---mirroring the restrict-solve-prolongate structure of block-spin methods.
Whether this analogy extends to the full renormalization group structure (fixed points, universality, semi-group composition~\citep{wilson1971renormalization}) is an open question for future work.

This structural parallel suggests a broader connection.
``Hierarchical feature extraction'' in deep learning---the progressive coarsening performed by ResNets, U-Nets, and transformer stages---bears a resemblance to renormalization group flow, where each layer integrates out short-range correlations and passes effective degrees of freedom to the next scale~\citep{he2019mgnet}.
Metriplector makes this connection more concrete: $\psifield$ dynamics provides the field evolution, $T^{\mu\nu}$ provides the feature readout, and $\rho$ provides the coarse-graining operator.
A unified architecture that combines the full $T^{\mu\nu}$ readout (from CIFAR and Reacher), iterative equilibrium solving (from Sudoku), multi-scale coarse-graining (from the object layer), and port-Hamiltonian action conditioning (from Reacher) is a natural convergence point.

\paragraph{Limitations and future work.}
CIFAR-100 accuracy (81.03\%) trails DenseNet-BC (82.8\%, 25.6M) despite using 10$\times$ fewer parameters.
Scaling to ImageNet will test whether the $K$-field bottleneck remains sufficient.
For language modeling, the causal Poisson model is more sample-efficient (3.6$\times$ fewer tokens) but less compute-efficient ($\sim$2$\times$ more GPU-minutes, $\sim$7$\times$ slower per-GPU throughput) than GPT, and our evaluation uses 512-token sequences vs.\ GPT's 1024.
The PG leaderboard SOTA (1.119 BPB) exceeds both models using orthogonal techniques.
For Reacher, 88\% CEM success slightly exceeds the strongest baseline (LeWM, 86\%) at 16$\times$ fewer parameters; with 50 evaluation episodes, the 2-percentage-point margin corresponds to a single episode and should be interpreted as statistically comparable rather than a definitive surpass.
Extension to higher-dimensional control tasks (e.g., Cheetah, Walker) and contact-rich manipulation will test the scalability of the port-Hamiltonian action conditioning and symplectic dynamics beyond the 6-DOF Reacher domain.

\section{Related Work}
\label{sec:related}

\paragraph{Attention, transformers, and state space models.}
The attention mechanism~\citep{bahdanau2015attention}, generalized by the transformer~\citep{vaswani2017attention}, computes $\softmax(QK^\top/\sqrt{d})V$---a single-step weighted average on a complete graph that provides universal function approximation~\citep{yun2020transformers}.
Vision Transformers~\citep{dosovitskiy2020image}, Swin~\citep{liu2021swin}, and LLMs~\citep{brown2020language,devlin2019bert,touvron2023llama} demonstrate the effectiveness of this approach at scale.
State space models---S4~\citep{gu2022efficiently}, Mamba~\citep{gu2024mamba}, and Mamba-2~\citep{dao2024mamba2}---reformulate sequence modeling as linear recurrences, achieving $O(N)$ complexity; Mamba-2 further showed that structured SSMs are equivalent to a form of structured attention.
RWKV~\citep{peng2023rwkv} takes a similar linear-recurrence approach.
Our causal Poisson scan is also a linear recurrence ($O(N\log N)$ via parallel associative scan), but derives from the screened Poisson equation rather than signal processing or attention decomposition.

\paragraph{Energy-based models.}
Hopfield networks~\citep{hopfield1982neural} showed that associative memory can be implemented as energy minimization.
LeCun~\citep{lecun2006tutorial} generalized this to Energy-Based Models (EBMs) for prediction.
Boltzmann machines~\citep{hinton1984boltzmann,ackley1985learning} and their restricted variants~\citep{smolensky1986information} extended energy-based learning to generative modeling.
Modern Hopfield networks~\citep{ramsauer2021hopfield} connect energy-based retrieval to transformer attention.
Equilibrium Propagation~\citep{scellier2017equilibrium} bridges EBMs and backpropagation by computing gradients through equilibrium states---related to our CG-based equilibrium approach, though we solve the equilibrium exactly rather than approximating it.
Metriplector uses both branches of the GENERIC equation, applying the coupled Hamiltonian-dissipative formulation to constraint satisfaction, image recognition, and language modeling.

\paragraph{Diffusion, flow, and score-based models.}
Diffusion models~\citep{ho2020denoising,song2021score} and their continuous-time formulation as score-based SDEs~\citep{song2021score} generate data by reversing a noise process governed by Langevin dynamics---a stochastic PDE where the learned score $\nabla_x \log p_t(x)$ plays a role analogous to a source term.
Flow matching~\citep{lipman2023flow} and rectified flows~\citep{liu2023flow} simplify this to deterministic ODE transport, learning a velocity field that maps noise to data.
Poisson Flow Generative Models~\citep{xu2022poisson} use the Poisson equation specifically, connecting to our screened Poisson solver.
These approaches share Metriplector's premise that PDE dynamics can serve as the computational mechanism, but apply it to generation (sampling from learned distributions) rather than discriminative tasks (classification, constraint satisfaction, dynamics prediction).
The key structural difference is that diffusion and flow models learn a \emph{time-varying} vector field over data space, while Metriplector learns \emph{spatial operators} (conductance, Poisson bracket, source) that define a fixed PDE whose solution provides the readout.

\paragraph{Hamiltonian and metriplectic learning.}
Neural ODEs~\citep{chen2018neural} introduced continuous-depth networks via ODE solvers.
Hamiltonian Neural Networks~\citep{greydanus2019hamiltonian} learn energy functions whose symplectic dynamics model physical systems; Symplectic ODE-Net~\citep{zhong2020symplectic} enforces structure via symplectic integrators.
Lagrangian Neural Networks~\citep{cranmer2020lagrangian} use the Euler-Lagrange formulation.
Hamiltonian Generative Networks~\citep{toth2020hamiltonian} demonstrated that Hamiltonian structure can be leveraged for generative modeling from high-dimensional observations without restrictive domain assumptions.
Port-Hamiltonian Neural Networks~\citep{desai2021port} and structure-preserving networks~\citep{galimberti2023hamiltonian} guarantee non-vanishing gradients; Port-Hamiltonian Deep Graph Networks~\citep{heilig2025phdgn} achieve state-of-the-art among MPNNs on the LRGB Peptides benchmark (AP\,=\,0.70) by leveraging energy conservation for long-range propagation.
Building on the ODE interpretation of residual networks~\citep{haber2017stable}, Chang et~al.~\citep{chang2018reversible} showed that reversible (Hamiltonian-inspired) ODE architectures achieve competitive accuracy on CIFAR-10/100; Rusch and Mishra~\citep{rusch2021unicornn} used Hamiltonian oscillator dynamics for RNNs, achieving 98.4\% on permuted sequential MNIST.
Noether Networks~\citep{alet2021noether} meta-learn conserved quantities as regularizers for sequential prediction; Noether's Razor~\citep{vanderouderaa2024noether} extends this by learning Noether symmetries for Hamiltonian systems.
Metriplector uses Noether's theorem differently---to derive the readout features ($T^{\mu\nu}$) from evolved fields rather than as a conservation regularizer.
Sosanya and Greydanus~\citep{sosanya2022dissipative} decompose dynamics into a Hamiltonian and a Rayleigh dissipation function via an implicit Helmholtz decomposition, demonstrated on damped mass-spring systems (2D state) and NOAA ocean current data.
The GENERIC framework~\citep{grmela1997dynamics,ottinger2005beyond} provides a general framework for non-equilibrium dynamics via $\dot{z} = L\nabla E + M\nabla S$.
Morrison~\citep{morrison1984bracket,morrison1986paradigm} introduced the metriplectic bracket---combining antisymmetric (Poisson) and symmetric (metric) components---and later established the Hamiltonian formulation for ideal fluids~\citep{morrison1998hamiltonian}; Olver~\citep{olver1993applications} connects Lie group symmetries to conservation laws.

\paragraph{Learning metriplectic dynamics.}
A substantial body of work learns metriplectic (GENERIC) structure from data for scientific simulation, operating on low-dimensional physical state trajectories.
Hern\'andez et~al.~\citep{hernandez2021structure} introduced structure-preserving neural networks (SPNN) that learn $E$, $S$, $L$, and $M$ while guaranteeing energy conservation and entropy production, validated on a double pendulum (4D state) and Couette flow of an Oldroyd-B viscoelastic fluid; they later extended this to latent-variable discovery via sparse autoencoders for thermodynamics-aware reduced-order models of high-dimensional FEM simulations~\citep{hernandez2021deep}.
Lee et~al.~\citep{lee2021machine} proposed a novel parameterization of dissipative brackets (GNODE) with exact preservation of Casimir invariants and the fluctuation-dissipation theorem, validated on two gas containers exchanging heat and volume (4D) and a thermoelastic double pendulum (10D), later extended in~\citep{lee2022learning}.
Zhang et~al.~\citep{zhang2022gfinns} developed GFINNs with modular architectures enforcing GENERIC degeneracy conditions on the same benchmarks plus Langevin dynamics (3D), and proved a universal approximation theorem.
Gruber et~al.~\citep{gruber2023energetically,gruber2025efficiently} developed Neural Metriplectic Systems (NMS) with quadratic scaling in state dimension and universal approximation guarantees (ICLR~2025), achieving 0.01 MSE on the gas container benchmark versus 0.12 for standard Neural ODEs.
Extensions include port-metriplectic neural networks enabling modular learning of open thermomechanical subsystems~\citep{hernandez2023portmetriplectic}, thermodynamics-informed graph neural networks achieving ${<}3\%$ relative error on Couette flow and bending beams on irregular meshes~\citep{hernandez2022thermodynamics} with a local formulation scaling to 9,000-particle SPH simulations~\citep{hernandez2024local}, data-driven coarse-grained particle dynamics for star polymers and colloidal suspensions from high-speed video~\citep{hernandez2025particles}, metriplectic conditional flow matching for generative modeling of dissipative dynamics~\citep{baheri2025mcfm}, and meta-learning frameworks for few-shot generalization across parametric families of GENERIC systems~\citep{jing2025metalearning}.
Gruber, Lee, and Trask~\citep{gruber2023brackets} applied bracket-based dynamics to deep graph neural networks for node classification on Cora, CiteSeer, and PubMed, reporting that their partially dissipative ``double bracket'' variant (83\% on Cora) outperformed the full metriplectic formulation (57--70\%) and was competitive with GAT.
Metriplector operates on learned latent representations rather than observed physical states, uses input-dependent operator configuration rather than fixed learned brackets, and extracts features via the stress-energy tensor $T^{\mu\nu}$.
The hard-constraint parameterizations for degeneracy conditions ($M\nabla E = 0$, $L\nabla S = 0$) developed in the metriplectic learning literature are complementary and could strengthen future Metriplector variants.

\paragraph{Physics-inspired architecture primitives.}
CliffordNet~\citep{ji2026cliffordnet} applies Clifford geometric products to neural features, computing symmetric (inner product) and antisymmetric (wedge product) interactions via an efficient sparse rolling mechanism.
This decomposition is related to the symmetric-antisymmetric split underlying the metric and Poisson brackets in Metriplector, though CliffordNet derives from algebraic completeness of the geometric product while Metriplector derives from non-equilibrium thermodynamics.
Kolmogorov-Arnold Networks (KAN)~\citep{liu2024kan} replace fixed activation functions with learned univariate splines on graph edges, grounding architecture design in the Kolmogorov-Arnold representation theorem; Metriplector instead grounds its primitive in field-theoretic dynamics and Noether's theorem.
Geometric algebra foundations~\citep{doran2003geometric,hestenes1984clifford} provide the mathematical framework shared by several of these approaches.
Group-equivariant CNNs~\citep{cohen2016group}, E(2)-steerable networks~\citep{weiler2019general}, E(n) equivariant GNNs~\citep{satorras2021en}, tensor field networks~\citep{thomas2018tensor}, and SE(3)-equivariant models~\citep{du2022se3} explore related symmetry-preserving architectures.

\paragraph{World models and latent dynamics.}
World models learn environment dynamics for model-based planning and control.
Ha \& Schmidhuber~\citep{ha2018world} combined VAE encoders with RNN dynamics; Dreamer~\citep{hafner2020dream} and DreamerV3~\citep{hafner2023mastering} achieve strong results via latent imagination with learned value functions.
The Joint Embedding Predictive Architecture (JEPA)~\citep{lecun2022path} advocates prediction in representation space rather than pixel space, avoiding the cost of generative decoding.
LeWM~\citep{maes2024lewm} instantiates JEPA with a ViT-Tiny encoder and causal transformer predictor, demonstrating stable training without pre-trained encoders via SIGReg regularization.
DINO-WM~\citep{zhou2024dinowm} uses frozen DINOv2 features as the representation space.
PLDM~\citep{guo2024pldm} pre-trains an encoder before learning latent dynamics.
TD-MPC2~\citep{hansen2024tdmpc2} combines world modeling with temporal difference learning for scalable control.
Metriplector's Reacher instantiation follows the JEPA paradigm---predicting in latent space without pixel reconstruction---but replaces transformer-based dynamics with metriplectic PDE evolution, canonical symplectic coupling, port-Hamiltonian action conditioning, and stress-energy readout, achieving comparable CEM success (88\%) with 16$\times$ fewer parameters.

\paragraph{Equilibrium models.}
Deep Equilibrium Models~\citep{bai2019deep} find fixed points of implicit layers.
Monotone Operator Equilibrium Networks~\citep{winston2020monotone} guarantee existence and uniqueness.
Our Poisson solve is a single-step equilibrium computation with guaranteed convergence (SPD system matrix), avoiding the iterative fixed-point search that DEQs require.

\paragraph{Differentiable optimization layers.}
OptNet~\citep{amos2017optnet} embeds quadratic programs in neural networks.
Differentiable convex optimization layers~\citep{agrawal2019differentiable} generalize this.
Vlastelica et~al.~\citep{vlastelica2020differentiation} differentiate through blackbox combinatorial solvers.
Our CG solver with adjoint implicit differentiation~\citep{hestenes1952methods,shewchuk1994introduction} follows this tradition but exploits SPD structure for efficient $O(N)$ memory gradients.

\paragraph{Physics-informed and physics-simulation networks.}
PINNs~\citep{raissi2019physics,karniadakis2021physics} encode PDE residuals as soft constraints.
Neural operators~\citep{li2021fourier} and DeepONet~\citep{lu2021deeponet} learn mappings between function spaces.
Graph Network Simulators~\citep{sanchez2020learning,pfaff2021learning} and message-passing PDE solvers~\citep{brandstetter2022message} learn physics from data on meshes.
In contrast to soft-constraint approaches, the Metriplector CG solver finds the exact PDE equilibrium, which guarantees physical consistency of the solution.

\paragraph{Graph neural networks and spectral methods.}
Message Passing Neural Networks~\citep{gilmer2017neural,scarselli2009graph,battaglia2018relational} provide a general framework for learned computation on graphs; see~\citet{bronstein2021geometric2} and~\citet{wu2020comprehensive} for surveys.
Spectral approaches---ChebNet~\citep{defferrard2016convolutional}, spectral CNNs~\citep{bruna2014spectral}---operate on graph Laplacian eigenvectors, connecting to spectral graph theory~\citep{chung1997spectral,spielman2007spectral}.
GCN~\citep{kipf2017semi}, GAT~\citep{velickovic2018graph}, and DeepGCNs~\citep{li2019deepgcns} learn spatial aggregation rules.
Xu et~al.~\citep{xu2019powerful} characterized GNN expressiveness via the Weisfeiler-Leman hierarchy.
Topping et~al.~\citep{topping2022understanding} showed that over-squashing in GNNs relates to graph curvature; Alon \& Yahav~\citep{alon2021bottleneck} identified the information bottleneck.
The Poisson solve computes global equilibrium in a single pass; information propagation range is controlled by the damping parameter $\lambda$ rather than by message-passing depth, which relates to the over-squashing phenomenon identified by~\citet{topping2022understanding} and~\citet{alon2021bottleneck}.

\paragraph{Multigrid methods and learned pooling.}
Classical multigrid~\citep{briggs2000multigrid,trottenberg2001multigrid,hackbusch1985multi} accelerates PDE solvers via fine-coarse cycling.
He \& Xu~\citep{he2019mgnet} showed that CNNs can be interpreted as multigrid iterations.
U-Net~\citep{ronneberger2015unet} implements a learned V-cycle.
DiffPool~\citep{ying2018diffpool} introduced learned soft-assignment graph pooling, where a GNN predicts cluster assignments and pools node features accordingly.
Our object layer follows this restrict-solve-prolongate pattern, with the distinction that the coarse-level computation is a Poisson solve rather than further message passing.

\paragraph{Neural Sudoku and constraint satisfaction.}
Sudoku is NP-complete to solve in general~\citep{yato2003complexity}, though the standard 9$\times$9 case is tractable; Felgenhauer \& Jarvis~\citep{felgenhauer2006enumerating} enumerate valid grids.
Recurrent Relational Networks~\citep{palm2018recurrent} solve Sudoku via message passing on a constraint graph ($>$96\% board accuracy).
SATNet~\citep{wang2019satnet} embeds a differentiable MAXSAT solver; \citet{zhang2021sudoku} explore neural logic machines.
All inject constraint structure (row, column, box edges or MAXSAT solvers) into the architecture.
Metriplector operates on a minimal 8-connected spatial grid without injecting constraint structure.

\paragraph{Neural algorithmic reasoning.}
Veli\v{c}kovi\'c \& Blundell~\citep{velickovic2021neural} formalize neural algorithmic reasoning; the CLRS benchmark~\citep{velivckovic2022clrs} evaluates neural networks on classical algorithms.
GNN approaches~\citep{ibarz2022generalist,tang2020towards} can learn algorithms but struggle with out-of-distribution sizes.
On maze pathfinding, the $\lambda/N$ scaling of the graph Laplacian enables generalization from 15$\times$15 training grids to 39$\times$39 evaluation grids (F1\,=\,0.95--0.99).

\section{Conclusion}
\label{sec:conclusion}

The stress-energy tensor $T^{\mu\nu}$, the conserved quantity of spatial translation symmetry, provides an effective and principled readout from learned field dynamics---this is the central finding of this work.
Built on this observation, Metriplector instantiates the metriplectic framework as a neural architecture primitive across five domains:
F1\,=\,1.0 on maze pathfinding (43.8K parameters) with robust size generalization;
97.2\% exact Sudoku solve rate (120K parameters) with zero structural injection;
81.03\% on CIFAR-100 (2.26M parameters);
1.182 BPB on language modeling with 3.6$\times$ fewer training tokens than a GPT baseline;
and 88\% CEM success on Reacher robotic control with 16$\times$ fewer parameters than transformer baselines.

Four design principles underpin these results and transfer across domains:
(1) the \emph{operator-from-input} separation, where the representation $\mathbf{h}$ configures the energy landscape and the fields $\psifield$ evolve on it;
(2) the \emph{metriplectic spectrum}, where task complexity determines which channels of the dynamics to activate---from pure dissipation (Poisson equation) to full Hamiltonian-dissipative coupling;
(3) the \emph{Noether readout}, where the stress-energy tensor provides the feature extraction from evolved fields;
(4) \emph{port-Hamiltonian action conditioning}, where the action modulates PDE operators rather than the state, enabling the same physics to model controllable systems with stable autoregressive rollouts;
(5) \emph{canonical symplectic structure}, where fields are organized as conjugate $(q,p)$ pairs with a fixed Poisson bracket $J_\Omega$, grounding the dynamics in Hamiltonian mechanics;
(6) \emph{multi-step autoregressive training}, where scheduled warmup from teacher forcing to self-prediction closes the exposure bias gap between training and evaluation.

The architecture has not been tested at transformer scale, and significant open questions remain---notably whether symplectic integrators can stabilize deeper dynamics, and whether a single architecture can span all five domains.
The consistent parameter efficiency across diverse tasks spanning reasoning, recognition, language, and control, however, indicates that physical structure encoded at the level of the architecture primitive can substantially reduce the capacity needed to solve a problem.


\bibliographystyle{plainnat}

\appendix

\section{Implicit Differentiation}
\label{app:implicit_diff}

Given $\psifield^* = A^{-1}\mathbf{b}$ where $A = L_W + \Lambda$ and downstream loss $\Lcal$:
\begin{align}
    \frac{\partial \Lcal}{\partial \mathbf{b}} &= A^{-1} \frac{\partial \Lcal}{\partial \psifield^*} = \mathbf{v}, \label{eq:grad_source} \\
    \frac{\partial \Lcal}{\partial w_{ij}} &= -\mathbf{v}_i(\psifield^*_i - \psifield^*_j) - \mathbf{v}_j(\psifield^*_j - \psifield^*_i), \label{eq:grad_w}
\end{align}
where $\mathbf{v} = A^{-1}(\partial \Lcal / \partial \psifield^*)$ is the adjoint variable.
The same CG solver is reused for both forward and adjoint solves, requiring $O(N)$ total memory.

\section{Dirichlet Energy Derivation}
\label{app:dirichlet}

Setting $\nabla_{\psifield} \Ecal_{\text{Dir}} = 0$ from Eq.~\eqref{eq:dirichlet}:
\begin{equation}
    \sum_{j \in \Ncal(i)} w_{ij}(\psifield_i - \psifield_j) + \Lambda_i \psifield_i = b_i \implies (L_W + \Lambda)\psifield = \mathbf{b}.
\end{equation}
Since $L_W + \Lambda$ is SPD (Proposition~\ref{prop:wellposed}), this is the unique global minimum.

\section{Lie Algebra of the Poisson Bracket}
\label{app:lie_algebra}

The Poisson bracket on the space of observables defines a Lie algebra.
For the $K$-field system with Poisson tensor $J_{ab}$, the bracket of two observables $F(\psifield)$ and $G(\psifield)$ is:
\begin{equation}
    \{F, G\}_J = \sum_{a,b} \frac{\partial F}{\partial \psifield_a} J_{ab} \frac{\partial G}{\partial \psifield_b}.
\end{equation}

\paragraph{Skew-symmetry.}
Since $J_{ab} = -J_{ba}$, we have $\{F, G\} = -\{G, F\}$ directly.

\paragraph{Jacobi identity.}
For the finite-dimensional Poisson tensor used in Metriplector ($J$ is a learned skew-symmetric matrix, constant in $\psifield$), the Jacobi identity $\{F, \{G, H\}\} + \{G, \{H, F\}\} + \{H, \{F, G\}\} = 0$ holds automatically because $J$ has zero curvature (it is a constant-coefficient Poisson structure).
For $\psifield$-dependent $J$, the Jacobi identity imposes non-trivial constraints on the dependence~\citep{marsden1999introduction}.
In the current architecture, $J$ is parameterised as a learned skew-symmetric matrix (independent of $\psifield$), so the Jacobi identity is satisfied by construction.

\paragraph{From Casimirs to Noether symmetries.}
Classical Hamiltonian mechanics distinguishes Casimir invariants---functions $C$ satisfying $\{C, F\} = 0$ for \emph{all} observables $F$---from energy conservation along a specific flow.
Casimirs are fixed by the bracket structure at initialisation and do not emerge from learning.
For Metriplector, the more relevant conservation framework is \emph{Noether's theorem}: every continuous symmetry of the learned energy functional $E(\psifield)$ has a corresponding conserved current.
Unlike Casimirs, Noether symmetries emerge from the learned energy landscape, change as the model trains (new symmetries discovered, old ones broken), and represent genuinely learned abstract structure.
The stress-energy tensor $T^{\mu\nu}$ that serves as Metriplector's readout is precisely the Noether current collection associated with translational invariance of the field dynamics (Appendix~\ref{app:noether_currents}).
The architecture has been monitoring these emergent conserved quantities since its first training step.

\section{Noether Currents for the Stress-Energy Readout}
\label{app:noether_currents}

We derive the stress-energy tensor readout from Noether's theorem for a system of $K$ scalar fields on a 2D domain.

\paragraph{Lagrangian density.}
Consider the Lagrangian density:
\begin{equation}
    \mathcal{L} = \frac{1}{2}\sum_{a=1}^{K} \left[(\partial_x \psifield_a)^2 + (\partial_y \psifield_a)^2\right] - V(\psifield_1, \ldots, \psifield_K),
\end{equation}
which describes $K$ interacting scalar fields with gradient energy.

\paragraph{Translation invariance.}
Under a spatial translation $x^\mu \to x^\mu + \epsilon^\mu$ (with $\delta\psifield_a = 0$), the Noether current is the canonical stress-energy tensor:
\begin{equation}
    T^{\mu\nu} = \sum_{a=1}^{K} \frac{\partial \mathcal{L}}{\partial (\partial_\mu \psifield_a)} \partial^\nu \psifield_a - \eta^{\mu\nu}\mathcal{L} = \sum_{a=1}^{K} \partial^\mu \psifield_a \cdot \partial^\nu \psifield_a - \eta^{\mu\nu}\mathcal{L}.
\end{equation}
The off-diagonal spatial components give the cross-field stress:
\begin{equation}
    T^{xy}_{ab} = \partial_x \psifield_a \cdot \partial_y \psifield_b.
\end{equation}

\paragraph{Decomposition into symmetric and antisymmetric parts.}
The tensor $T^{xy}_{ab}$ decomposes as:
\begin{align}
    E_{ab} &= \tfrac{1}{2}(T^{xy}_{ab} + T^{xy}_{ba}) = \tfrac{1}{2}(\partial_x \psifield_a \partial_y \psifield_b + \partial_x \psifield_b \partial_y \psifield_a), \quad \text{(energy density)} \\
    V_{ab} &= \tfrac{1}{2}(T^{xy}_{ab} - T^{xy}_{ba}) = \tfrac{1}{2}(\partial_x \psifield_a \partial_y \psifield_b - \partial_x \psifield_b \partial_y \psifield_a). \quad \text{(vorticity)}
\end{align}
The symmetric part $E_{ab}$ measures co-located edge structure; the antisymmetric part $V_{ab}$ measures rotational structure (corners, junctions).
Together they yield $K^2$ independent features: $K$ diagonal + $\binom{K}{2}$ cross-energy + $\binom{K}{2}$ vorticity.

\paragraph{Additional Noether currents.}
Rotational invariance ($x \to x\cos\theta - y\sin\theta$, $y \to x\sin\theta + y\cos\theta$) yields angular momentum:
\begin{equation}
    L_a = x \cdot p_{y,a} - y \cdot p_{x,a}, \qquad p_{\mu,a} = \psifield_a \cdot \partial_\mu \psifield_a.
\end{equation}
Scale invariance ($x \to \lambda x$, $y \to \lambda y$) yields the dilation current:
\begin{equation}
    D_a = x \cdot p_{x,a} + y \cdot p_{y,a}.
\end{equation}
These position-dependent currents provide physics-native positional encoding without learned position embeddings.

\section{GENERIC Degeneracy Conditions}
\label{app:degeneracy}

The GENERIC equation $\dot{z} = L\nabla E + M\nabla S$ requires two degeneracy conditions that prevent the two brackets from interfering:

\paragraph{Condition 1: $M \cdot \nabla E = 0$.}
The dissipative bracket cannot change the total energy.
In Metriplector's dissipative branch, this is satisfied by construction: the Onsager matrix $M = L_W + \Lambda$ generates dynamics $\dot{\psifield} = -M\psifield + \mathbf{b}$ that minimise the Dirichlet energy $\Ecal = \frac{1}{2}\psifield^\top M \psifield - \mathbf{b}^\top\psifield$.
At equilibrium ($\dot{\psifield} = 0$), the energy is at its unique minimum.
During evolution, $dE/dt = -\|\dot{\psifield}\|^2_M \leq 0$: energy is dissipated, not created.

\paragraph{Condition 2: $L \cdot \nabla S = 0$.}
The Hamiltonian bracket cannot produce entropy.
For the Poisson bracket $J$ with energy $E(\psifield) = \frac{1}{2}\sum_a |\psifield_a|^2$, the Hamiltonian flow $\dot{\psifield}_a = \sum_b J_{ab}\psifield_b$ satisfies $dE/dt = \psifield^\top J \psifield = 0$ by skew-symmetry of $J$.
The entropy functional $S(\psifield)$ (typically the negative Dirichlet energy in the dissipative branch) is likewise preserved: $dS/dt = \nabla S^\top J \nabla S = 0$.

\paragraph{Soft degeneracy in practice.}
In the full Metriplector architecture, the degeneracy conditions are not enforced exactly in each layer.
Instead, they emerge approximately through training: the gradient signal rewards configurations where the Poisson bracket preserves important features (approximate energy conservation) and the dissipative bracket drives convergence (entropy increase).
The architectural separation---$J$ is skew-symmetric by construction, $M$ is SPD by construction---guarantees the \emph{structural} requirements ($J^\top = -J$, $M \succeq 0$) exactly, while the \emph{functional} degeneracy ($M\nabla E = 0$, $L\nabla S = 0$) is approximately learned.

\end{document}